\documentclass[10pt]{article}

\usepackage[preprint]{rlj}

\usepackage{amssymb}
\usepackage{mathtools}
\usepackage{mathrsfs}
\usepackage{graphicx}
\usepackage{subcaption}
\usepackage[space]{grffile}
\usepackage{url}
\usepackage{lipsum}

\usepackage{bm}
\usepackage{caption}
\usepackage{booktabs}
\usepackage{enumitem}
\usepackage{makecell}
\usepackage{microtype}
\usepackage{multirow}
\usepackage{tikz}
\usepackage{wrapfig}
\usepackage{xcolor}

\DeclareMathOperator{\argmax}{argmax}

\newcommand{\multirowcmidruleheight}{-0.5\dimexpr\aboverulesep+\belowrulesep+\cmidrulewidth}

\usetikzlibrary{automata,snakes,positioning}
\usetikzlibrary{positioning,shapes,matrix,calligraphy}

\newcommand{\myjustify}[1]{{#1\parfillskip=0pt\par}}
\renewcommand{\paragraph}[1]{\textbf{#1}\hspace{10pt}}

\title{Disentangling Recognition and Decision Regrets\\in Image-Based Reinforcement Learning}
\setrunningtitle{Disentangling Recognition and Decision Regrets in Image-Based Reinforcement Learning}

\author{%
    Alihan H\"uy\"uk\textsuperscript{\normalfont 1$\dagger$},
    A.\ Ryo Koblitz\textsuperscript{\normalfont 2},
    Atefeh Mohajeri\textsuperscript{\normalfont 2},
    Matthew Andrews\textsuperscript{\normalfont 2}}
\emails{}
\affiliations{%
    \textsuperscript{1}\textbf{Harvard University,} \textsuperscript{2}\textbf{Nokia Bell Labs} \\
    \textsuperscript{$\dagger$}Work done as an intern at Nokia Bell Labs}


\begin{document}

\maketitle

\begin{abstract}
    \looseness-1
    In image-based reinforcement learning (RL), policies usually operate in two steps: first extracting lower-dimensional features from raw images (the ``recognition'' step), and then taking actions based on the extracted features (the ``decision'' step). Extracting features that are spuriously correlated with performance or irrelevant for decision-making can lead to poor generalization performance, known as \textit{observational overfitting} in image-based RL. In such cases, it can be hard to quantify how much of the error can be attributed to poor feature extraction vs.\ poor decision-making. To disentangle the two sources of error, we introduce the notions of \textit{recognition regret} and \textit{decision regret}. Using these notions, we characterize and disambiguate the two distinct causes behind observational overfitting: \textit{over-specific representations}, which include features that are not needed for optimal decision-making (leading to high decision regret), vs.\ \textit{under-specific representations}, which only include a limited set of features that were spuriously correlated with performance during training (leading to high recognition regret). Finally, we provide illustrative examples of observational overfitting due to both over-specific and under-specific representations in maze environments and the Atari game Pong.
\end{abstract}

\section{Introduction}

\begin{wrapfigure}[9]{r}{.56\linewidth}
    \vspace{-\baselineskip-6pt}%
    \definecolor{peach}{rgb}{1.0, 0.8549019607843137, 0.7254901960784313}%
    \definecolor{lavender}{rgb}{0.9019607843137255, 0.9019607843137255, 0.9803921568627451}%
    \tikzset{
        block/.style = {draw, fill=white, rectangle, minimum height=3em, minimum width=3em},
        trapz/.style = {draw,
            fill=peach,
            trapezium,
            shape border rotate=270,
            minimum height=20mm,
            minimum width=20mm,
            trapezium left angle=80,
            trapezium right angle=80,
            rounded corners=1},
        trapezium stretches=true,
        latent/.style = {
            rectangle split,
            rectangle split every empty part={},
            rectangle split empty part width=0.5mm,
            rectangle split empty part height=0.5mm,
            rectangle split parts=5,
            rectangle split,
            draw=black,
            rounded corners=1,
            rectangle split part fill=peach,
            rectangle split draw splits=true},
        head/.style={
            rectangle,
            rounded corners=1,
            minimum height=8mm,
            minimum width = 20mm,
            draw=black,
            fill=lavender}}%
    \resizebox{\linewidth}{!}{%
        \begin{tikzpicture}[auto, >=latex, pin distance=10mm, thick]
    
            \node (obs) {\includegraphics[clip,trim={16px 16px 16px 16px},height=19mm]{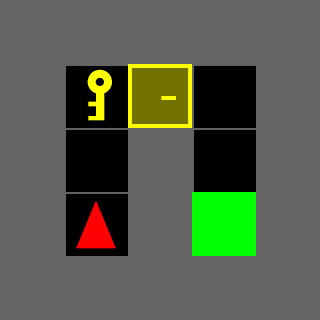}};
            \node[on grid, align=center,below=11mm of obs,anchor=north] (obslab) {Observation\\[-1pt]$x_t\sim\omega(s_t)$};
        
            \node[trapz, right=9mm of obs] (enc) {$\rho'\in\Delta(Z)^{X}$};
            \node[on grid, align=center,below=11mm of enc,anchor=north] (enclab) {Encoder};
        
            \node[latent, right=12mm of enc] (lat) {};
            \node[on grid, align=center, below=11mm of lat,anchor=north] {Representation\\[-1pt]$z_t\sim\rho'(x_t)$};
            \node[head, right=12mm of lat] (head) {$\pi\in\Delta(A)^{Z}$};
            \node[on grid, align=center, below=11mm of head,anchor=north] {Output Head};
            \node[circle,fill=lavender,draw=black,right=9mm of head] (act){$a_t$};
            \node[align=center, on grid, below=11mm of act,anchor=north] {Action\\[-1pt]$a_t\sim\pi(z_t)$};
        
            \draw(obs) edge[->] (enc);
            \draw(enc) edge[->] (lat);
            \draw(lat) edge[->] (head);
            \draw(head) edge[->] (act);
        
            \draw [line width=1.25pt,decorate,decoration={calligraphic brace,amplitude=10pt,raise=12mm}]
          (enc.west) -- (head.east) node[midway,yshift=15.5mm]{Agent};
        
        \end{tikzpicture}}%
    \vspace{-6pt}%
    \caption{\myjustify{
        Block diagram of an agent consisting of a \textit{recognition policy} (encoder) followed by a \textit{decision policy} (output head).}}%
    \label{fig:diagram}%
\end{wrapfigure}
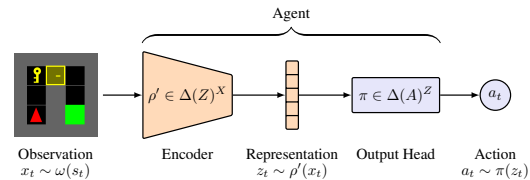

\myjustify{
    In image-based reinforcement learning (RL), it is common for agents to act in two steps by following two sub-policies one after another \citep[e.g.\ the popular RL package Stable Baselines3,][]{raffin2021stable}: (i)~First, a \textit{recognition policy} generates representations of input images by extracting lower-dimensional features from high-dimensional pixels. (ii)~Then, a \textit{decision policy} generates actions based on those representations (see Figure~\ref{fig:diagram}). Naturally, if an agent acting in this way is performing poorly, this would be partly due to poor feature extraction and partly due to poor decision-making, in other words, partly because the recognition policy is suboptimal and partly because the decision policy is suboptimal. Here, identifying which policy is the main source of error in an underperforming RL system can provide crucial information for debugging and improving that system. However, quantifying the individual contribution of each policy to the overall error can be challenging, especially when agents are trained in an end-to-end fashion and both policies are optimized jointly.}

Our goal in this paper is to disentangle these two potential sources of error, namely poor recognition vs.\ poor de\-ci\-sion-making. As a motivating example of how such a distinction can play an important role in building RL systems, consider \textit{observational overfitting} typically encountered in image-based RL \citep{song2020observational}: Image-based policies often do not generalize well to unseen scenarios because they fail to extract \textit{critical features} that need to be identified in images to be able to make good decisions. Instead, they end up extracting either \textit{irrelevant features} that should not matter when making decisions or \textit{spurious features} that were mistakenly correlated with performance during training. As an example of irrelevant features, a self-driving car might encode in its representations the makes and models of other cars in the traffic although good driving does not require recognizing car makes and models, and consequently, might struggle when encountering a new car with a never-before-seen make and model. As an example of spurious features, a self-navigating robot might memorize landmarks in the background as cues for when to perform certain maneuvers rather than identifying actual obstacles, and consequently, might fail to navigate a new course when the same landmarks are no longer there. In order to avoid observational overfitting, we should be able to tell when a recognition policy fails to extract critical features over irrelevant or spurious features.

We achieve our goal of disentangling the error induced by an agent's recognition policy vs.\ their decision policy by introducing the notions of \textit{recognition regret} and \textit{decision regret}. Conventionally, \textit{regret} measures the difference between an agent's performance and the best possible performance of any agent. Our regret notions break this conventional definition into two parts: (i) First, relying on representations generated by a given recognition policy as input to decision policies, rather than the original images directly, might already reduce the best performance that can possibly be achieved by any decision policy---the difference is the recognition regret. (ii) Then, a given decision policy might still fall short of achieving the best possible performance of any decision policy---the difference is the decision regret. In other words, the recognition regret measures how much the representations of a given recognition policy bottlenecks the down-stream performance of decision policies while the decision regret measures how much potential performance is missed out by a given decision policy.

Making use of our definitions, we characterize and disambiguate two distinct causes behind observational overfitting: \textit{over-specificity} or \textit{under-specificity} of representations generated by recognition policies. Over-specific representations include ``too many'' features, often irrelevant ones in addition to critical ones. Since these representations still include the critical features, they do not necessarily limit the downstream performance of decision policies and hence do not incur much recognition regret. However, the presence of irrelevant features might cause decision policies to treat images with similar critical features as if they are unrelated when they do not share the same irrelevant features, which would lead to high decision regret. In contrast, under-specific representations include ``too few'' features, usually only capturing spurious indicators of performance during training. This might cause decision policies to treat unrelated images as if they are similar to each other just because they feature the same spurious indicators, which would lead to high recognition regret over decision regret.

\paragraph{Contributions}
Our contributions are three-fold: First, in \textbf{Section~\ref{sec:contribution1}}, we define recognition and decision regrets, which disentangle the regret induced by poor recognition policies vs.\ poor decision policies. Next, in \textbf{Section~\ref{sec:contribution2}}, we apply these definitions to the problem setting of generalization in RL. Through recognition and decision regrets in terms of generalization performance, we characterize over-specific and under-specific representations as two distinct modes of observational overfitting in image-based RL. Finally, in \textbf{Section~\ref{sec:contribution3}}, we provide illustrative examples of observational overfitting due to both over-specific and under-specific representations in maze environments as well as the Atari game Pong. These examples demonstrate that the notions of recognition and decision regrets can be powerful tools in articulating the cause of observational overfitting in image-based RL applications.

\vspace{-5pt}
\section{Related Work}
\vspace{-5pt}

\paragraph{Representations in Image-Based RL}
In RL, algorithms for training policies in the form of neural networks generally fall into one of two categories: action-value fitting methods such as \citet{mnih2015human} and policy gradient methods such as \citet{mnih2016asynchronous,schulman2017proximal}. When these neural networks take images as input, one of the most common architectures used in practice is an encoder consisting of convolutional layers followed by either an action-value head if using an action-value fitting method \citep[e.g.][]{mnih2015human,raffin2021stable}, or followed by a policy head for generating actions and a value head for estimating values if using a policy gradient method \citep[e.g.][]{cobbe2019quantifying,raffin2021stable}. With this type of architecture (an encoder followed by output heads), the intermediary output of the encoder can be viewed as a lower-dimensional representation of the input images. While a lot of previous work has focused on learning ``better'' representations for various purposes \citep[e.g.][]{sonar2021invariant,raileanu2021decoupling,dabney2021value,agarwal2021contrastive,stooke2021decoupling,zhang2022efficient,eysenbach2022contrastive}, our focus in this paper is to \textit{quantify} the quality of an encoder's representations as far as its contribution to the end performance is concerned. We capture the performance loss due to a poor encoder (i.e.\ a poor recognition policy) through recognition regret. Meanwhile, any additional losses that might be incurred by the subsequent action-value/policy/value heads (i.e.\ the decision policies) are captured through decision regret.

\looseness-1
Others have also asked what constitutes a ``good'' representation in RL: According to \citet{bellemare2019geometric,le2022generalization}, representations should be sufficient in approximating the value of all stationary policies. Meanwhile, \citet{ghosh2020representations,wang2024investigating} survey different representation learning schemes and evaluate a variety of properties such as stability/robustness, capacity, and redundancy. Different from all this work, we consider the representations extracted by a recognition policy not in isolation but in relation to a specific decision policies, which may have been trained jointly with the recognition policy or subsequently (i.e.\ recognition regret w.r.t.\ decision regret). According to our work, representations should not only limit performance (cf.\ recognition regret) but also facilitate the training of high-performing decision policies (cf.\ decision regret). Under-specific representations fail at the former while over-specific representations fail at the latter.

\paragraph{Generalization in RL \& Observational Overfitting}
Training policies that generalize beyond their particular training conditions remain a significant challenge in RL \citep{zhang2018dissection,packer2018assessing,nichol2018gotta}. This is partly because policies may overfit due to a multitude of confounded factors that can be hard to disambiguate from each other. For instance, some of these factors may include the characteristics of the training environment like whether it is deterministic or not \citep{bellemare2013arcade,pinto2017robust,rajeswaran2017towards,machado2018revisiting}, the method of exploration during training like how many initial states are sampled \citep{zhang2018dissection,zhang2018study}, or even the hyper-parameters of the training algorithms like the discount factor $\gamma$ \citep{jiang2015dependence}.

We focus our attention on observational overfitting specifically as characterized by \citep{song2020observational} because of its close relationship to learned representations in image-based RL. This particular type of overfitting occurs in cases where different environments are all semantically the same with the same state spaces and the same transition dynamics but generate different images/observations. For instance, a game might have the same underlying game mechanics in all of its levels but each level might have its own unique art style \citep[e.g.\ the popular generalization benchmarks][]{nichol2018gotta,cobbe2019quantifying}. In such cases, if a policy fails to learn representations that accurately identify the shared states between different environments, and instead extracts features that are merely cosmetic and differ from one environment to another, then that policy might fail to generalize to new environments that have unseen observation dynamics \citep{sonar2021invariant}. In this paper, we go one step beyond \citet{song2020observational} and describe two distinct modes of observational overfitting: due to over-specificity, where the policy successfully extracts the true state but also features that are irrelevant to transition dynamics, and due to under-specificity, where the policy fails to extract the true state in the first place and ends up relying on spurious features that do not generalize.

\section{Defining Recognition and Decision Regrets}
\label{sec:contribution1}

In this section, we give formal definitions of decision regret and recognition regret. We also aim to provide an intuitive understanding of these definitions through a worked example in Section~\ref{sec:contribution1a} and empirical examples in Section~\ref{sec:contribution1b}. Later in Section~\ref{sec:contribution2}, we will apply these two concepts to the problem of generalization and describe two potential causes behind observational overfitting: over-specific vs.\ under-specific representations. Finally in Section~\ref{sec:contribution3}, we will provide illustrative examples of both these causes in action.

\paragraph{Environments}
Consider partially-observable environments: $\varepsilon=(\sigma,\tau,\omega,r)$ with \textit{state} space~$S$, \textit{action} space $A$, and \textit{observation} space $X$, where $\sigma\in\Delta(S)$ is the initial state distribution, $\tau\in\Delta(S)^{S\times A}$ is the \textit{transition dynamics}, $\omega\in\Delta(X)^{S}$ is the \textit{observation dynamics}, and $r\in\Delta(\mathbb{R})^{S}$ is the \textit{reward dynamics}. Beginning from an initial state $s_1\sim\sigma$, at each time step $t\in\{1,2,\ldots\}$, the environment emits an observation $x_t\sim\omega(s_t)$ based on the current state, gives out a reward $r_t\sim r(s_t)$, and finally transitions into a new state $s_{t+1}\sim\tau(s_t,a_t)$ based on the actions of an agent.

\looseness-1
\paragraph{Recognition \& Decision Policies}
For some \textit{representation} space  $Z$, we denote with $\rho\in\Delta(Z)^{Z\times X}$ \textit{recognition policies} and with $\pi\in\Delta(A)^{Z}$ \textit{decision policies}. Beginning from an initial representation $z_0\in Z$, an agent following these policies first generates an updated representation $z_t\sim\rho(z_{t-1},x_t)$ based on the most recent observation according to the recognition policy and then takes an action $a_t\sim\pi(z_t)$ based on the updated representation according to the decision policy. For agents with no memory, the recognition policy would be a function of the most recent observation only (i.e.\ an encoder) such that $z_t\sim\rho(z_{t-1},x_t)\doteq\rho'(x_t)$ where $\rho'\in\Delta(Z)^{X}$. Given a pair of recognition and decision policies $(\rho_0,\pi_0)$ and a discount factor $\gamma\in(0,1)$, their (discounted) \textit{value} in environment~$\varepsilon$ is
\begin{align}
    V_{\varepsilon}(\rho_0,\pi_0) = \mathbb{E}_{\varepsilon,\rho_0,\pi_0}\left[{\textstyle\sum_t} \gamma^tr_t\right]
\end{align}
\paragraph{Recognition \& Decision Regrets}
Having defined value, we can now quantify the suboptimality of a given policy $(\rho_0,\pi_0)$ in environment $\varepsilon$ through its \textit{regret}, which is simply the gap between its value and the \textit{optimal value} $V_{\varepsilon}^*$ (i.e.\ how much of the attainable value is missed out by the given policy):
\begin{align}
    R_{\varepsilon}(\rho_0,\pi_0) &\doteq V_{\varepsilon}^* - V_{\varepsilon}(\rho_0,\pi_0) \doteq \max\nolimits_{\rho,\pi}V_{\varepsilon}(\rho,\pi) -  V_{\varepsilon}(\rho_0,\pi_0)
\end{align}
We decompose this regret into two parts, \textit{recognition regret} and \textit{decision regret} respectively:
\begin{align}
    R_{\varepsilon}^{\text{rec}}(\rho_0) ~&\doteq~ \max\nolimits_{\rho,\pi} V_{\varepsilon}(\rho,\pi) - \max\nolimits_{\pi}V_{\varepsilon}(\rho_0,\pi) \\
    R_{\varepsilon}^{\text{dec}}(\rho_0,\pi_0) ~&\doteq~ \max\nolimits_{\pi}V_{\varepsilon}(\rho_0,\pi) - V_{\varepsilon}(\rho_0,\pi_0)
\end{align}
such that $R=R^{\text{rec}}+R^{\text{dec}}$. Intuitively, the recognition regret measures how much $\rho_0$ alone limits the value attainable by subsequent decision policies. While generally, the value can be as high as the optimal value $V^*=\max_{\rho,\pi}V(\rho,\pi)$, when conditioned on $\rho_0$ specifically, it becomes bounded by $\max_{\pi}V(\rho_0,\pi)\leq V^*$ (and the difference is defined as $R^{\text{rec}}$). Meanwhile, the decision regret measures how much of the value still attainable after conditioning on $\rho_0$ is missed out by $\pi_0$. Although the value can still be as high as $\max_{\pi}V
(\rho_0,\pi)$ under $\rho_0$, pairing $\rho_0$ with $\pi_0$ leads only to a value of $V(\rho_0,\pi_0)$ (and the difference is defined as $R^{\text{dec}}$).

\subsection{A Worked Example}
\label{sec:contribution1a}

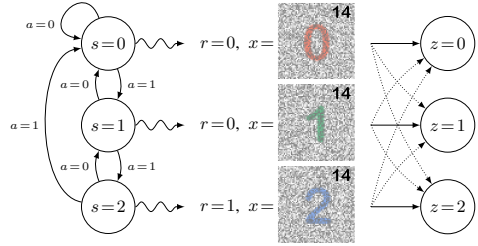
\begin{wrapfigure}[11]{r}{.48\linewidth}
    \centering%
    \vspace{-\baselineskip}%
    \vspace*{6pt}%
    \hspace*{21pt}%
    \resizebox{.8\linewidth}{!}{%
        \begin{tikzpicture}[auto, >=latex,font=\small, pin distance=10mm]
            \tikzstyle{every pin edge}=[->, shorten <=0pt, snake=snake, line before snake=0pt, line after snake=4pt]
            
            \node[state] (s0) {$s\!=\!0$};
            \node[overlay,right=9mm of s0] (o0) {~~$r\!=\!0$,~~$x\!=\!\!\!$\makecell{\includegraphics[width=18mm]{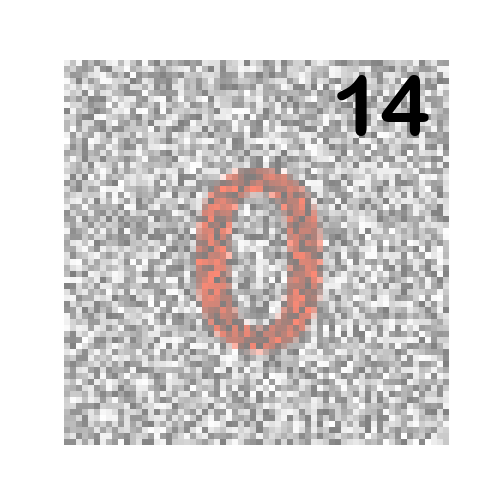}}};
            \draw[->, shorten <=0pt, snake=snake, line before snake=0pt, line after snake=4pt] (s0) -> (o0);
            \node[state, right=9mm of o0] (z0) {$z\!=\!0$};
    
            \node[state, below=5mm of s0] (s1) {$s\!=\!1$};
            \node[right=9mm of s1] (o1) {~~$r\!=\!0$,~~$x\!=\!\!\!$\makecell{\includegraphics[width=18mm]{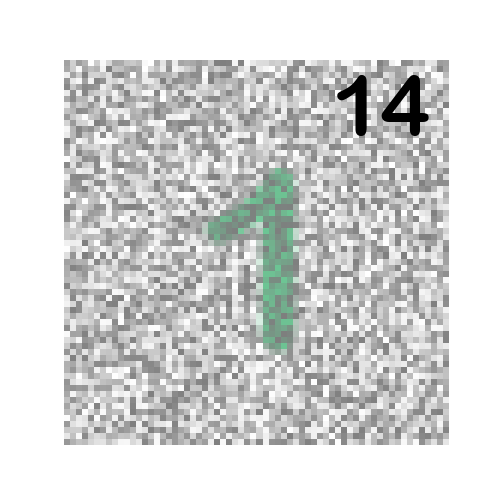}}};
            \draw[->, shorten <=0pt, snake=snake, line before snake=0pt, line after snake=4pt] (s1) -> (o1);
            \node[state, right=9mm of o1] (z1) {$z\!=\!1$};
        
            \node[state, below=5mm of s1] (s2) {$s\!=\!2$};
            \node[overlay,right=9mm of s2] (o2) {~~$r\!=\!1$,~~$x\!=\!\!\!$\makecell{\includegraphics[width=18mm]{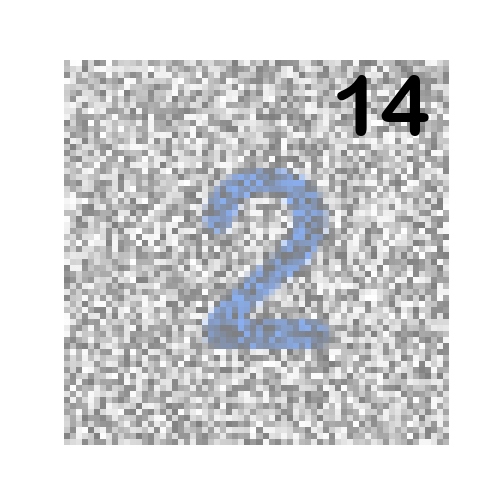}}};
            \draw[->, shorten <=0pt, snake=snake, line before snake=0pt, line after snake=4pt] (s2) -> (o2);
            \node[state, right=9mm of o2] (z2) {$z\!=\!2$};
    
            \tikzset{every loop/.style={in=170,out=110,min distance=8mm,looseness=0.8}}
            \draw[->]   (s0) edge[overlay,loop above, left] node[overlay,font=\tiny,xshift=-3.5pt,yshift=-7.5pt]{$a\!=\!0$} (s0);
            \draw[->]   (s0) edge[bend left=20, right] node[font=\tiny] {$a\!=\!1$} (s1);
            \draw[->]   (s1) edge[bend left=20, left] node[overlay,font=\tiny] {$a\!=\!0$} (s0);
            \draw[->]   (s1) edge[bend left=20, right] node[font=\tiny] {$a\!=\!1$} (s2);
            \draw[->]   (s2) edge[bend left=20, left] node[overlay,font=\tiny] {$a\!=\!0$} (s1);
            \draw[->]   (s2) edge[overlay,bend left=80, left,looseness=0.8] node[overlay,font=\tiny] {$a\!=\!1$} (s0);

            \draw[->](o0) edge (z0);
            \draw[densely dotted] (o0.east) edge[->, bend right=10] (z1);
            \draw[densely dotted] (o0.east) edge[->, bend right=10] (z2);
        
            \draw[->](o1) edge (z1);
            \draw[densely dotted] (o1.east) edge[->, bend left=15] (z0);
            \draw[densely dotted] (o1.east) edge[->, bend right=15] (z2);
        
            \draw[->](o2) edge (z2);
            \draw[densely dotted] (o2.east) edge[->, bend left=10] (z1);
            \draw[densely dotted] (o2.east) edge[->, bend left=10] (z0);
        \end{tikzpicture}}%
    \caption{States \textbf{(left)}, observations \textbf{(center)}, and representations \textbf{(right)} for our 3-state example at $t\!=\!14$.}%
    \label{fig:worked_example}
\end{wrapfigure}

To illustrate our definitions, we consider the simple 3-state environment with $S=\{0,1,2\}$ shown in Figure~\ref{fig:worked_example}. Only state~$2$ produces a reward so the optimal policy takes action~$1$ in states~$0$ and $1$, and action~$0$ in state~$2$, since this ensures we achieve reward~$1$ in half of the time steps. If the initial state is $0$, the optimal value in this environment is given by $v_0$ in the equations $v_0=\gamma v_1$, $v_1=\gamma v_2$, $v_2=1+\gamma v_1$, that is  $V^* = \gamma^2+\gamma^4+\cdots$, which equals $4.263$ when the discount factor is $\gamma=0.9$.
The observation space $X$ is a union of image spaces $\cup_{i\in S,t\in\{1,2,\ldots\}} I_{it}$, where $I_{it}$ is a set of noisy representations of the number $i$, with a clear black representation of the time $t$ in the top right corner. The colors of the integers in $I_{0t},I_{1t},I_{2t}$ are red, green, blue respectively. When $s_t=i$, observation $x_t$ is an image in $I_{it}$ selected randomly.

\myjustify{
    \paragraph{The Recognition Policy}
    The recognition policy $\rho$ is a function of the most recent observation only and is defined by a mapping $\rho'(x)$ from each image $x$ to the representation space $Z=\{0,1,2\}$ (hence $Z=S$). The optimal recognition policy ``should'' recognize the depiction of $s_t$ in the image (and ignore the time in the top right corner). However, the noise in the images means that $I_{0t},I_{1t},I_{2t}$ are hard to distinguish and so consider a suboptimal policy $\rho'_0$ defined for some error parameter $\delta$. When $s_t=i$, with probability $1-\delta$, $\rho'_0(x_t)=i$ (i.e.\ the recognition policy is correct), and with probability $\delta$, $\rho'_0(x_t)$ is uniformly distributed across $Z$ (i.e.\ the recognition policy makes an arbitrary choice).}

When $\delta$ is small, the optimal decision policy is still to take action $1$ in states $0$ and $1$, and action $0$ in state $2$ (since we always want to get to state $2$ as soon as possible). However, with probability~$\delta$, $\rho'_0(x_t)$ is a random element from $Z$ due to incorrect recognition, and when that happens the action is $1$ with probability $2/3$ and $0$ with probability $1/3$. The value in the environment is now given by $v_0$ in the equations $v_0=\gamma v_0(\delta/3)+\gamma v_1(1-\delta/3)$, $v_1=\gamma v_0(\delta/3)+\gamma v_2(1-\delta/3)$, $v_2 = 1+\gamma v_0(2\delta/3)+\gamma v_1(1-2\delta/3)$. When $\gamma=0.9$ and $\delta=0.1$, the solution to these equations is $(v_0,v_1,v_2)=(3.992,4.451,4.978)$ hence the \mbox{\textit{recognition regret}} $R^{\text{rec}}(\rho_0) =4.263-3.992=0.271$.

\paragraph{The Decision Policy}
Suppose we now couple this suboptimal recognition policy with a $\delta$-optimal decision policy $\pi_0$, meaning one that based on the state recognition takes the optimal action with probability $1-\delta$ and a random action with probability $\delta$. If we combine the recognition and decision errors, if the true state is $0$ then $\pi_0$ takes the wrong action~$0$ with probability $\delta/3+\delta/2$, if the true state is $1$ then $\pi_0$ takes the wrong action~$0$ with probability $\delta/3+\delta/2$, and if the true state is $2$ then $\pi_0$ takes the wrong action~$1$ with probability $2\delta/3+\delta/2$.
The value in the environment is now given by $v_0$ in the equations $v_0=\gamma v_0(\delta/3+\delta/2)+\gamma v_1(1-\delta/3-\delta/2)$, $v_1=\gamma v_0(\delta/3+\delta/2)+\gamma v_2(1-\delta/3+\delta/2)$, $v_2 = 1+\gamma v_0(2\delta/3+\delta/2)+\gamma v_1(1-2\delta/3-\delta/2)$. When $\gamma=0.9$ and $\delta=0.1$, the solution to these equations is $(v_0,v_1,v_2)=(3.680,4.126,4.666)$ hence the \mbox{\textit{decision regret}} $R^{\text{dec}}(\rho_0,\pi_0) = 3.992-3.680 = 0.312$.

\subsection{Empirical Examples}
\label{sec:contribution1b}

To illustrate our definitions further, we also provide empirical examples where we intentionally construct suboptimal decision policies and suboptimal recognition policies and highlight how the values of $R^{\text{rec}}$ and $R^{\text{dec}}$ differ between the two scenarios we construct.

\paragraph{Setup}
We consider two image-based environments: \textit{Minigrid}, which is a maze environment where the agent needs to navigate around walls to reach some target \citep{maxime2023minigrid}, and \textit{Pong}, which is an Atari game where the agent aims to score points against a computer opponent \citep{towers_gymnasium_2023}. 
In both environments, we consider policies with the same type of architecture, where the recognition policy, $z=\rho_{\text{CNN}}(x)$, is given by a convolutional neural network (CNN) and the decision policy, $a\sim\pi_{\text{MLP}}(z)$, is given by a multi-layer perceptron (MLP), and we train these policies using the PPO algorithm \citep{schulman2017proximal} from the skrl package \citep{serrano2023skrl}. When presenting results, we shift and scale rewards so that the regret is at most one. Details regarding the setup can be found in the supplementary material.

\paragraph{Suboptimal Decision Policies}
We construct agents with \textit{suboptimal} decision policies by restricting them to occasionally take random actions. First, we modify the network~$\pi_{\text{MLP}}$, denoting with $\tilde{\pi}$ the resulting decision policy, such that
\begin{align}
    a \sim \tilde{\pi}(z) = \Big\{\begin{matrix*}[l]
        \scriptstyle \text{Uniform}(A) &\scriptstyle\text{with probability}\quad p \\[-3pt]
        \scriptstyle \pi_{\text{MLP}}(z) &\scriptstyle\text{with probability}\quad 1-p
    \end{matrix*}
\end{align}
\looseness-1
where $p\in[0,1]$ is the probability of taking a forced random action and $\text{Uniform}(A)$ is the uniform distribution over
$A$. Then, we train the weights of the two networks, $\rho_{\text{CNN}}$ and $\pi_{\text{MLP}}$, as usual via PPO.

In this scenario, the occasional random actions would inevitably induce some regret $R(\rho_{\text{CNN}},\tilde{\pi})$. However, maximizing the rewards when actions are not forced to be taken randomly would still require finding a near-optimal $\rho_{\text{CNN}}$ and $\pi_{\text{MLP}}$ since non-random actions are still generated by $\rho_{\text{CNN}}\circ\pi_{\text{MLP}}$. Therefore, we would expect that $V(\rho_{\text{CNN}},\pi_{\text{MLP}})\approx V^*$ and $\pi_{\text{MLP}}\approx \argmax_{\pi}V(\rho_{\text{CNN}},\pi)$, and hence, $R^{\text{rec}}(\rho_{\text{CNN}}) = V^* - \max_{\pi}V(\rho_{\text{CNN}},\pi)\approx 0$. In other words, the occasional random actions should not induce much recognition regret but rather induce decision regret.

\begin{figure}[t]
    \centering
    \begin{subfigure}{.5\linewidth}
        \centering%
        \includegraphics[width=.5\linewidth,clip,trim={9pt 12pt 9pt 12pt}]{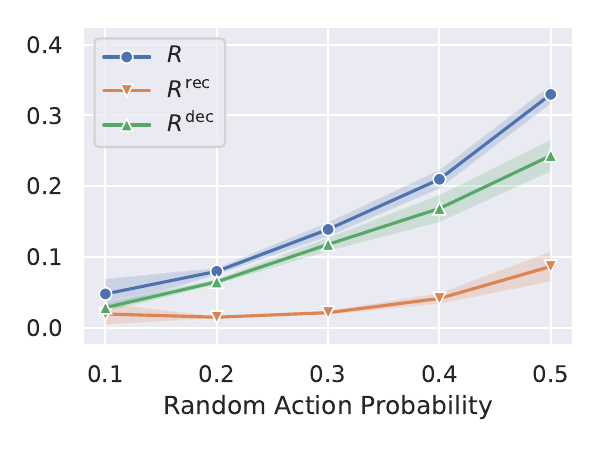}%
        \includegraphics[width=.5\linewidth,clip,trim={9pt 12pt 9pt 12pt}]{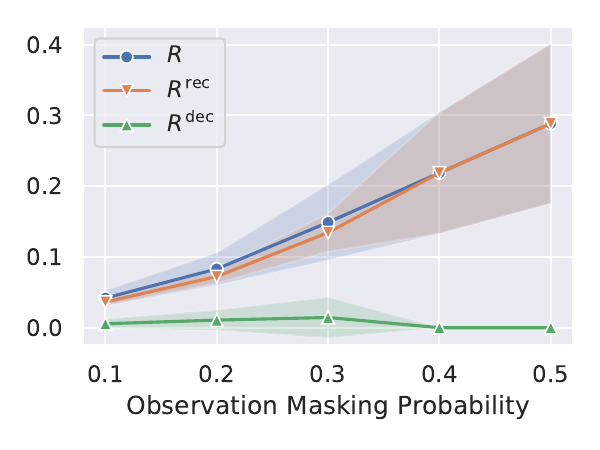}
        
        \vspace{-3pt}%
        \caption{\textbf{Minigrid}}
    \end{subfigure}%
    \begin{subfigure}{.5\linewidth}
        \centering%
        \includegraphics[width=.5\linewidth,clip,trim={9pt 12pt 9pt 12pt}]{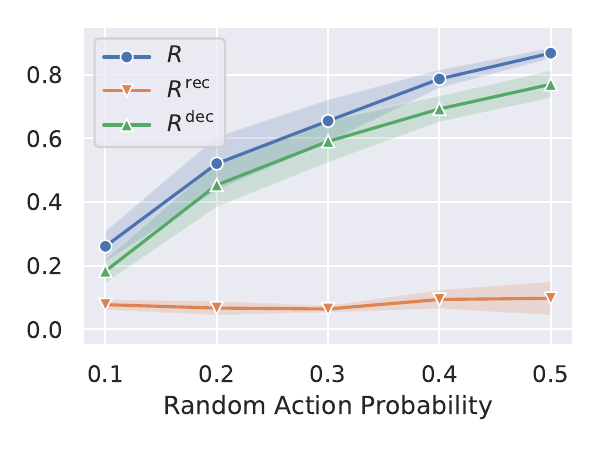}%
        \includegraphics[width=.5\linewidth,clip,trim={9pt 12pt 9pt 12pt}]{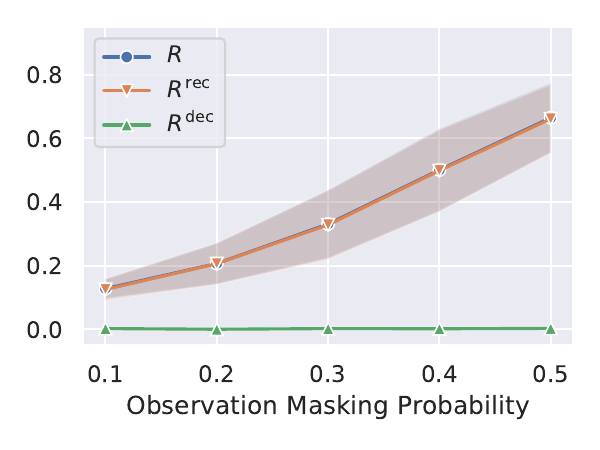}
        
        \vspace{-3pt}%
        \caption{\textbf{Pong}}
    \end{subfigure}

    \vspace{-6pt}%
    \caption{\myjustify{
        \textit{Recognition and Decision Regrets under Randomized Actions and Masked Observations.}
        Both scenarios lead to regret. For random actions, it is mostly decision regret, indicating: Even when forced to occasionally take random actions, agents still learn useful representations of their environment (to be able to optimize their non-random actions). For masked observations, it is recognition regret, indicating: Partially masked observations fundamentally limit the agent's ability to optimize their actions regardless of their decision policy.}}%
    \label{fig:regret}%
\end{figure}

Figure~\ref{fig:regret} confirms this intuition, where we plot $R$, $R^{\text{rec}}$, $R^{\text{dec}}$ of $(\rho_{\text{CNN}},\tilde{\pi})$ for varying $p$ on the left-side panels. Indeed, an increasing probability of random actions leads to increasing regret as well which is predominantly decision regret in both \textit{Minigrid} and \textit{Pong}. In these panels, plotting $R^{\text{rec}}$ and $R^{\text{dec}}$ requires estimating $\max_{\pi}V(\rho_{\text{CNN}},\pi)$, which we achieve by fixing the weights of $\rho_{\text{CNN}}$, setting $p=0$ to remove any restrictions on $\pi_{\text{MLP}}$ so that $\tilde{\pi}=\pi_{\text{MLP}}$, and re-training the weights of $\pi_{\text{MLP}}$.

\paragraph{Suboptimal Recognition Policies}
We construct agents with \textit{suboptimal} recognition policies by randomly masking parts of the images/observations before feeding them forward to an optimal agent. First, we train the weights of $\rho_{\text{CNN}}$ and $\pi_{\text{MLP}}$ without any modifications. Then, letting $X=\mathbb{R}^{W\times H\times C}$ be the space of $W$-by-$H$ images with $C$ channels, we modify the (pre-trained) network $\rho_{\text{CNN}}$, denoting with $\tilde{\rho}$ the resulting recognition policy, such that
\begin{align}
    z=\tilde{\rho}(x)=\rho_{\text{CNN}}(x \odot m),~~~ m_{ijk}=\Big\{\begin{matrix*}[l]
        \scriptstyle 0 &\scriptstyle\text{w/ prob.}~~~ p \\[-3pt]
        \scriptstyle 1 &\scriptstyle\text{w/ prob.}~~~ 1-p
    \end{matrix*}
\end{align}
where $p\in(0,1)$ is the probablity of masking an individual pixel-channel, $m\in \{0,1\}^{W\times H\times C}$ is a randomly sampled masking vector, and $\odot$ denotes element-wise multiplication.

In this scenario, the partially masked observations would induce some regret $R(\tilde{\rho},\pi_{\text{CNN}})$ as before. However this time, we would expect most of that regret to be recognition regret since switching from $\rho_{\text{CNN}}$ to $\tilde{\rho}$ would limit the performance of all decision policies $\pi$ not just $\pi_{\text{CNN}}$ specifically.

Again, Figure~\ref{fig:regret} confirms this intuition, where we plot $R$, $R^{\text{rec}}$, $R^{\text{dec}}$ of $(\tilde{\rho},\pi_{\text{MLP}})$ for varying $p$ on the right-side panels: An increasing masking probability leads to increasing regret that is predominantly recognition regret in both \textit{Minigrid} and \text{Pong}. Just like before, plotting $R^{\text{rec}}$ and $R^{\text{dec}}$ in these panels requires estimating $\max_{\pi}V(\tilde{\rho},\pi)$, which we achieve by fixing the weights of $\rho_{\text{CNN}}$ and the value of $p$ so that the recognition policy $\tilde{\rho}$ is completely fixed, and then retraining the weights of $\pi_{\text{MLP}}$.

\section{Practical Application:\hspace{\linewidth}Generalization in terms of Recognition and Decision Regrets}
\label{sec:contribution2}

In practice, it is not always possible to directly measure the decision, recognition, or overall regret of policies as these metrics require knowing optimal policies under various constraints. Often, our goal is to find such optimal policies in the first place. For instance, estimating $\max_{\pi}V(\rho_0,\pi)$ would require us to optimize a decision policy given a recognition policy. Suppose we happen to have a method to do so with the result being $\hat{\pi}^*\approx\argmax_{\pi}V(\rho_0,\pi)$. Then, we can estimate the decision regret as $\smash{R^{\text{dec}}}\approx V(\rho_0,\hat{\pi}^*)-V(\rho_0,\pi_0)$, which is indeed what we did in our empirical examples in Section~\ref{sec:contribution1b}. But in a practical application, the decision policy we are interested in analyzing, which we have been denoting as $\pi_0$, would most likely be the same policy as $\hat{\pi}^*$ in the first place, rendering our estimate meaningless as it would be equal to zero identically.

However, when measuring \textit{generalization performance}, that is the performance of policies optimized in some training environment but evaluated in some other test environment, then it can be possible to estimate recognition and decision regrets in a meaningful way (essentially by computing regret via policies optimized in the test environment to analyze policies that are optimized in the training environment). In this section, we develop this idea as a practical application of our definitions in Section~\ref{sec:contribution1}.
First, we introduce the notion of \textit{generalization regret} to quantify how much performance is missed out in the test environment due to suboptimal generalization from the training environment. Then, similar to our approach in the previous section, we decompose generalization regret into recognition and decision components. These two components allow us to characterize over-specific vs.\ under-specific representations, which we exemplify later in Section~\ref{sec:contribution3}.

\looseness-1
\paragraph{Generalization Regret}
Instead of a single environment~$\varepsilon$, we now consider two different environments: a \textit{training environment} $\varepsilon^{\text{train}}$ and a \textit{test environment} $\varepsilon^{\text{test}}$, which we denote together as $\mathcal{E}=(\varepsilon^{\text{train}},\varepsilon^{\text{test}})$. For instance, if we are interested in observational overfitting, these two environments might differ in terms of their observation dynamics $\omega$ (but not $\tau$).
Moreover, suppose that we are given a learning algorithm~$\mathcal{L}$ for training policies such that $\mathcal{L}(\varepsilon)\doteq(\mathcal{L}_{\rho}(\varepsilon),\mathcal{L}_{\pi}(\varepsilon))\approx\argmax_{\rho,\pi}V_{\varepsilon}(\rho,\pi)$ and $\mathcal{L}_{\pi}(\varepsilon|\rho_0)\approx\argmax_{\pi}V_{\varepsilon}(\rho_0,\pi)$.
Then, the generalization performance of policies trained in $\varepsilon^{\text{train}}$ via $\mathcal{L}$  but tested in $\varepsilon^{\text{test}}$ is given by $V_{\varepsilon^{\text{test}}}(\mathcal{L}(\varepsilon^{\text{train}}))$.
Similar to regret in Section~\ref{sec:contribution1}, we can consider this performance relative to the best possible test performance, which we call the \textit{generalization regret}:
\begin{align}
    {G\!R}{}_{\mathcal{E}}(\mathcal{L}) &= V^*_{\varepsilon^{\text{test}}} - V_{\varepsilon^{\text{test}}}(\mathcal{L}(\varepsilon^{\text{train}}))
\end{align}
Estimating this metric still requires knowing the optimal test policy $\argmax_{\rho,\pi}V_{\varepsilon^{\text{test}}}(\rho,\pi)$. However, unlike the case with regular regret, we can now obtain a meaningful estimate of $G\!R$ by using policies $\mathcal{L}(\varepsilon^{\text{test}})$ trained directly in the test environment as a proxy for the optimal test policy (since these policies are now distinctly different from the policies $\mathcal{L}(\varepsilon^{\text{train}})$ trained originally). This results in the following estimation of the generalization regret:
\begin{align}
    \widehat{G\!R}{}_{\mathcal{E}}(\mathcal{L}) = V_{\varepsilon^{\text{test}}}(\mathcal{L}(\varepsilon^{\text{test}})) - V_{\varepsilon^{\text{test}}}(\mathcal{L}(\varepsilon^{\text{train}}))
\end{align}
which measures how much the test performance could have been improved if it were to be possible to train policies directly in the test environment. Notice the subtle difference between this and $G\!R$: The latter measures how much of the test performance is missed out by training policies only in the training environment. When the policies trained in the training environment are already generalizable to the test environment, that is when $G\!R=0$, for sensible algorithms $\mathcal{L}$, $\smash{\widehat{G\!R}}$ should also be zero, meaning no performance gain should be possible by re-training policies in the test environment.

\paragraph{Relationship with Generalization Error}
Notably, the generalization regret is closely related to the \textit{generalization error} as previously defined by \citet{zhang2018dissection}:
\begin{align}
    {G\!E}{}_{\mathcal{E}}(\mathcal{L}) = V_{\varepsilon^{\text{train}}}(\mathcal{L}(\varepsilon^{\text{train}})) - V_{\varepsilon^{\text{test}}}(\mathcal{L}(\varepsilon^{\text{train}}))
\end{align}
which measures how much performance is retained when the policies trained in the training environment are rolled out in the test environment. Both definitions are ultimately concerned with the same generalization performance given by $V_{\varepsilon^{\text{test}}}(\mathcal{L}(\varepsilon^{\text{train}}))$. The difference is that, while $G\!E$ measures this performance relative to the training performance achieved in the training environment, $\smash{\widehat{G\!R}}$ measures it relative to the training performance that could have been achieved in the test environment. However, our definition $\smash{\widehat{G\!R}}$, being a type of regret, lends itself to the same decomposition as in Section~\ref{sec:contribution1} into recognition and decision components, which is not immediately possible with $G\!E$.

\paragraph{Recognition \& Decision Generalization Regrets}
Similar to Section 3, consider the intermediary case when only the decision policy has access to the test environment and the test value that would have been achieved then, which can be written as
\begin{align}
    V_{\varepsilon^{\text{test}}}(~~\rho=\mathcal{L}_{\rho}(\varepsilon^{\text{train}})~,~~\pi=\mathcal{L}_{\pi}(\varepsilon^{\text{test}}|\mathcal{L}_{\rho}(\varepsilon^{\text{train}}))~~) \label{eqn:intermediary}
\end{align}
Then, we decompose $\smash{\widehat{G\!R}}$ with respect to this intermediary value as follows:
\begin{align}
    \widehat{G\!R}{}^{\text{rec}}_{\mathcal{E}}(\mathcal{L}) &= V_{\varepsilon^{\text{test}}}(\mathcal{L}(\varepsilon^{\text{test}})) - \eqref{eqn:intermediary} \\
    \widehat{G\!R}{}^{\text{dec}}_{\mathcal{E}}(\mathcal{L}) &= \eqref{eqn:intermediary} - V_{\varepsilon^{\text{test}}}(\mathcal{L}(\varepsilon^{\text{train}}))
\end{align}
\looseness-1
Intuitively, $\smash{\widehat{G\!R}}{}^{\text{dec}}$ is the hypothetical performance that could be gained from re-training $\mathcal{L}_{\pi}(\varepsilon^{\text{train}})$ directly in the test environment, and $\smash{\widehat{G\!R}}{}^{\text{rec}}$ is the further performance gain of re-training $\mathcal{L}_{\rho}(\varepsilon^{\text{train}})$ in addition to $\mathcal{L}_{\pi}(\varepsilon^{\text{train}})$ (note that $\smash{\widehat{G\!R}}=\smash{\widehat{G\!R}}{}^{\text{rec}}+\smash{\widehat{G\!R}}{}^{\text{dec}}$ is the total gain).
A worked example, similar to the one in Section~\ref{sec:contribution1a}, can be found in the supplementary material.

\paragraph{Overfitting}
Finally, we are ready to characterize overfitting as well as different modes of overfitting using all the metrics we have defined so far, namely $R$, $\smash{\widehat{G\!R}}$, $\smash{\widehat{G\!R}{}^{\text{rec}}}$, and $\smash{\widehat{G\!R}{}^{\text{dec}}}$:
We say that policies $\rho_0,\pi_0 = \mathcal{L}(\varepsilon^{\text{train}})$ have \textit{overfitted} if the generalization regret~$\smash{G\!R}_{\mathcal{E}}(\mathcal{L})=R_{\varepsilon^{\text{test}}}(\mathcal{L}(\varepsilon^{\text{train}}))$ is high despite a relatively low training regret~$R_{\varepsilon^{\text{train}}}(\mathcal{L}(\varepsilon^{\text{train}}))$. This is said to be a case of \textit{observational overfitting} if the training and test environments share the same dynamics except for their observation dynamics such that $\varepsilon^{\text{train}}=(\sigma,\tau,\omega^{\text{train}},r)$ and $\varepsilon^{\text{test}}=(\sigma,\tau,\omega^{\text{test}},r)$. When overfitting occurs,
\begin{itemize}
    \item We say that representations generated by the recognition policy~$\rho_0$ are \textit{under-specific} if $\smash{\widehat{G\!R}{}^{\text{rec}}}$ is high but $\smash{\widehat{G\!R}{}^{\text{dec}}}$ is relatively low. Intuitively, under-specific representations include ``too few'' features which do not contain enough information to perform the task at hand (for instance, a blind recognition policy that outputs the same representation for all observations, $\rho'(x_t)=\emptyset$, or an overfitted recognition policy that only learned to extract spurious features). This limits the performance of all decision policies, leading to high $\smash{\widehat{G\!R}{}^{\text{rec}}}$.

    \item We say that representations generated by the recognition policy~$\rho_0$ are \textit{over-specific} if $\smash{\widehat{G\!R}{}^{\text{dec}}}$ is high but $\smash{\widehat{G\!R}{}^{\text{rec}}}$ is relatively low. Intuitively, over-specific representations include ``too many'' features which still contain enough information to perform the task at hand but also contain irrelevant information, for instance if the recognition policy has underfitted. This does not necessarily limit the performance that can be achieved by subsequent decision policies, leading to low $\smash{\widehat{G\!R}{}^{\text{rec}}}$, but it may cause a decision policy to take different actions in situations that are semantically the same (i.e.\ the same state~$s_t$) just because the irrelevant information differs (i.e.\ different representations~$z_t$), leading to high $\smash{\widehat{G\!R}{}^{\text{dec}}}$. Here, it is the decision policy that has overfitted, due to the over-specific representations generated by an underfitted recognition policy.
\end{itemize}

\section{Illustrative Examples}
\label{sec:contribution3}

\looseness-1
In this section, we give concrete examples of observational overfitting due to both under-specific representations and over-specific representations in two image-based environments: \textit{Minigrid} in Section~\ref{sec:contribution3a}, which is a collection of maze environments, and the Atari game \textit{Pong} in Section~\ref{sec:contribution3b}. These environments contrast each other well in terms of their scale.%
\footnote{Besides the scale contrast, these environments are commonly considered when exploring generalization in RL \citep[e.g.][]{sonar2021invariant,taiga2023investigating}. Moreover, they offer organically-generated images whereas the literature on observational overfitting tends to rely mostly on synthetically-generated images, including \citet{song2020observational}, which introduced the concept.}
\textit{Minigrid} outputs $7$-by-$7$ images with three channels corresponding to object types, object states, and color identifiers. Meanwhile, \textit{Pong}, after some post-processing, outputs $164$-by-$82$ images that are grayscale and we stack the four most recent images to account for memoryless agents.
In addition to these examples, we also highlight model selection as a potential use case of analyzing the specificity of representations in Section~\ref{sec:contribution3c}.

\begin{wrapfigure}[15]{r}{.36\linewidth}
    \vspace{-\baselineskip}%
    \begin{subfigure}{\linewidth}
        \includegraphics[width=42pt]{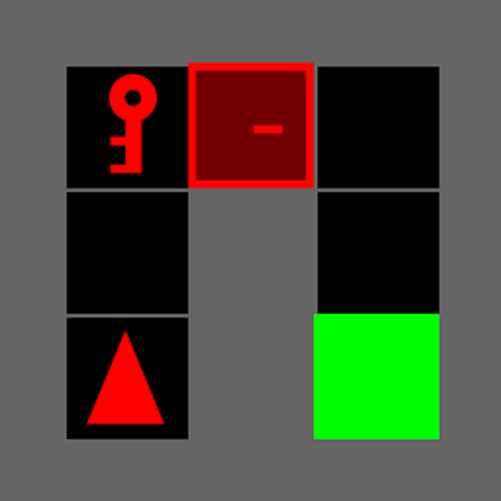}\hfill
        \includegraphics[width=42pt]{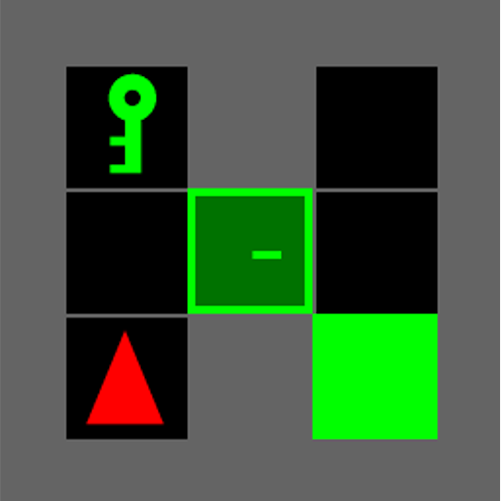}\hfill%
        \includegraphics[width=42pt]{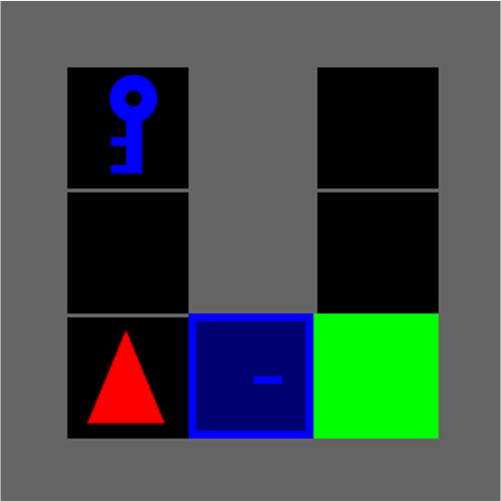}%
        \caption{\textbf{Training Configurations}}%
    \end{subfigure}%
    
    \vspace{3pt}%
    \begin{subfigure}{\linewidth}
        \includegraphics[width=42pt]{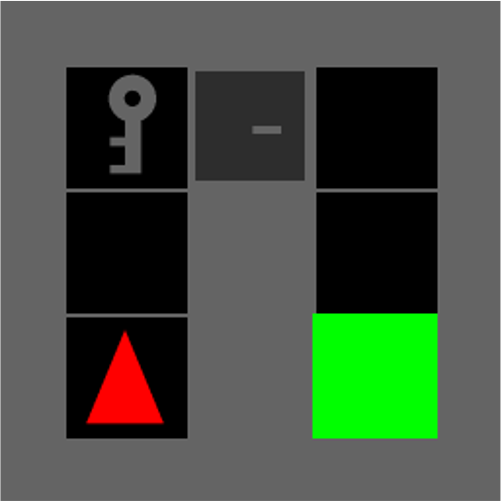}\hfill%
        \includegraphics[width=42pt]{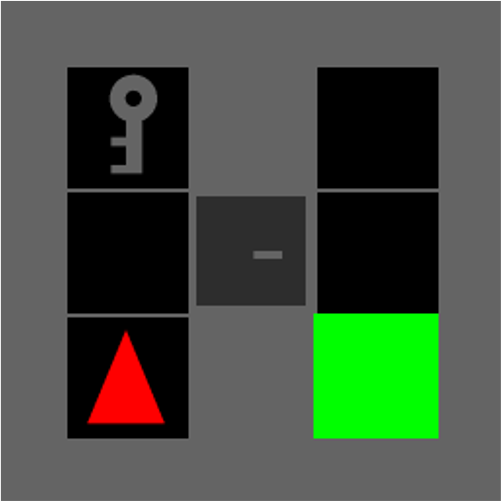}\hfill%
        \includegraphics[width=42pt]{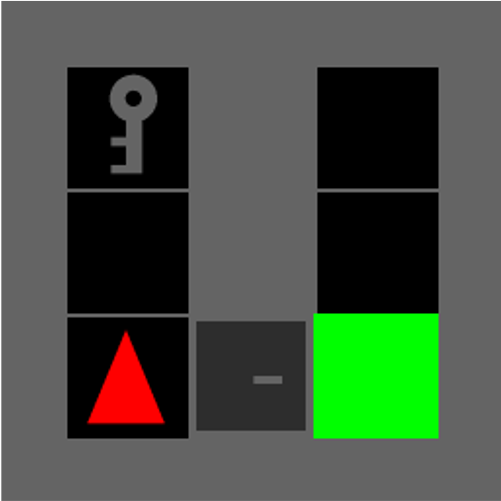}%
        \caption{\textbf{Test Configurations}}%
    \end{subfigure}%
    \vspace{-6pt}%
    \caption{\textit{Possible Maze Configurations.} The agent is the red triangle, and the goal is the green square. For training, the key-door color is determined based on the door location, but for testing, the key-door color is always gray (making it both an irrelevant and a spurious feature).}%
    \label{fig:minigrid-environments}%
\end{wrapfigure}

\looseness-1
As in Section~\ref{sec:contribution1b}, we train agents using the PPO algorithm. In all our experiments, the agents consist of a CNN that acts as the recognition policy followed by an MLP that acts as the decision policy (and followed by another parallel MLP that acts as the value head for the PPO algorithm). Details regarding the training setup can be found in the supplementary material.

\subsection{Examples in Maze Environments}
\label{sec:contribution3a}

\looseness-1
\paragraph{Setup}
We consider simple mazes given on a $3$-by-$3$ grid, where the agent needs to navigate to some goal square. We consider different maze configurations, but in each configuration, the goal is separated from the agent by a locked door, and to be able to reach the goal, the agent needs to first pick up a key and unlock the door. With uniform probability, the door can either be at (i) the \textit{North}-end,  (ii) the \textit{Center}, or (iii) the \textit{South}-end of the parting wall between the goal and the agent (see Figure~\ref{fig:minigrid-environments}).

To be able to create agents with over-specified or under-specified representations, our training and testing environments need to have \textit{irrelevant} and \textit{spurious} features. As the irrelevant feature, we introduce color: For the maze configurations in the training environment, the key-door pair is one of three colors: \textit{Red}, \textit{Green}, or \textit{Blue}. In the test environment, the key-door pair is always \textit{Gray}, which is never is never seen during training. These colors are purely visual and they have no impact on how the environment functions, making them an irrelevant feature to the completion of the maze. We turn the key-door color into a spurious feature as well by making it correlated with the door location during training: The north door is always red, the center door is always green, and the south door is always blue in the training configurations. Therefore, knowing  the key-door color alone is enough to be able to complete mazes by memorizing the solution to the maze corresponding to each color---but only during training (since the key-door color is always gray during testing).

\looseness-1
\paragraph{Agents}
We construct different agents by applying different \textit{pre-filters} to input images as part of the recognition policy before feeding them forward to the same network architecture. This technique allows us to freely control the properties of an agent's recognition policy and illustrate cases of over-specific vs.\ under-specific representations.
We primarily consider three agents/pre-filters (see the corresponding column in Table~\ref{tab:minigrid-results}, results for additional agents can be found in the supplementary material): (i)~\textit{Identity} makes no modifications to the input image. This means the key-door color is preserved as an irrelevant feature, making \textit{Identity} susceptible to learning over-specific representations. (ii)~\textit{HideDoor} hides the location of the true door by placing false doors in the maze but in the same color as the original door. This means \textit{HideDoor} can only rely on the key-door color as a spurious feature to navigate during training, which may lead to learning \textit{under-specific representations}. (iii)~\textit{HideColors} re-colors the key and the door as gray, the same color as the keys and the door in the test environment. This should make \textit{HideColors} easily generalizable to the test environment.

\begin{table}
    \centering
    \caption{All agents incur almost zero regret during training but only \textit{HideColors} generalizes well to the test environment. \textit{Identity} fails to generalize because its representations are sensitive to irrelevant features (notice the low similarity between the representations of mazes with the same door location but different key-door colors, especially between the unseen gray color and the others). This is characterized by high $\smash{\widehat{G\!R}}{}^{\text{dec}}$ (over-specificity). \textit{HideDoor} fails to generalize because its recognition policy has ``overfitted'' to extracting spurious indicators of optimal action trajectories (i.e.\ the key-door color). This is characterized by high $\smash{\widehat{G\!R}}{}^{\text{rec}}$ (under-specificity).}%
    \label{tab:minigrid-results}%
    \vspace{-6pt}%
    \resizebox{\linewidth-24pt}{!}{%
        \begin{tabular}{@{}lccc@{}c@{}c@{}}
            \toprule
            \multirow{2}{*}[\multirowcmidruleheight]{\bf Agent} & \multirow{2}{*}[\multirowcmidruleheight]{\bf Pre-Filter} & \multirow{2}{*}[\multirowcmidruleheight]{\bf Performance} & \multicolumn{3}{c@{\hspace{-10pt}}}{\bf Representation Similarity} \\
            \cmidrule(l){4-6}
            & & & All & By Color & {By Door Loc.} \\
            \midrule
            \makecell[l]{Identity\\(\bf\textit{Over-Specific})}
                & \makecell{\includegraphics[clip,trim={16px 16px 16px 16px},height=48pt]{figures/gen-minigrid/filter-id.png}}
                & \makecell{
                    $\mathmakebox[24pt][r]{R}=$~0.002 (0.003) \\
                    $\mathmakebox[24pt][r]{\smash{\widehat{G\!R}}}=$~0.911 (0.171) \\
                    $\mathmakebox[24pt][r]{\smash{\widehat{G\!R}}^{\text{rec}}}=$~0.264 (0.388) \\
                    $\mathmakebox[24pt][r]{\smash{\widehat{G\!R}}^{\text{dec}}}=$~0.647 (0.365)}
                & \raisebox{6pt}{\makecell{\includegraphics[height=64pt]{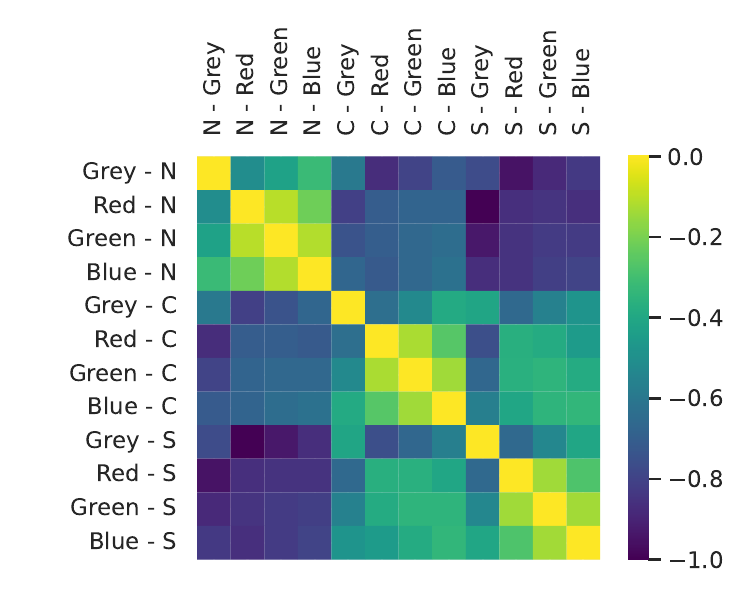}}}
                & \raisebox{1pt}{\makecell{\includegraphics[height=52pt]{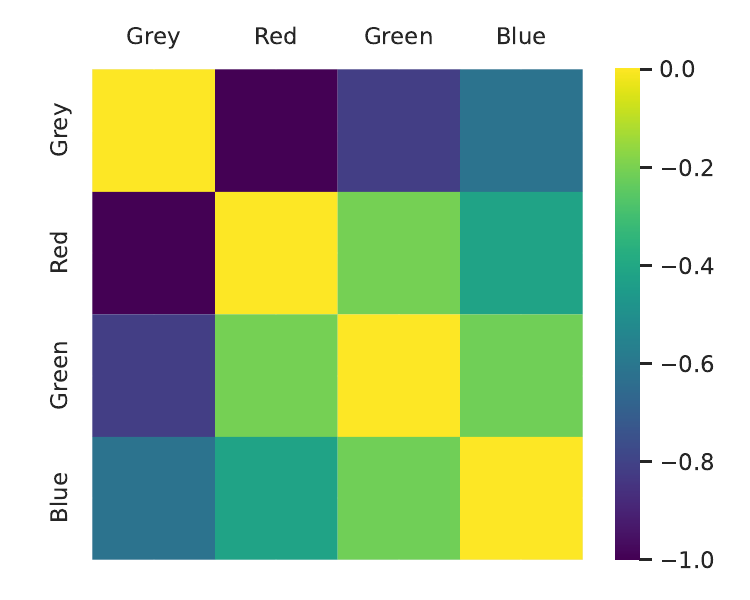}}}
                & \raisebox{1pt}{\makecell{\includegraphics[height=52pt]{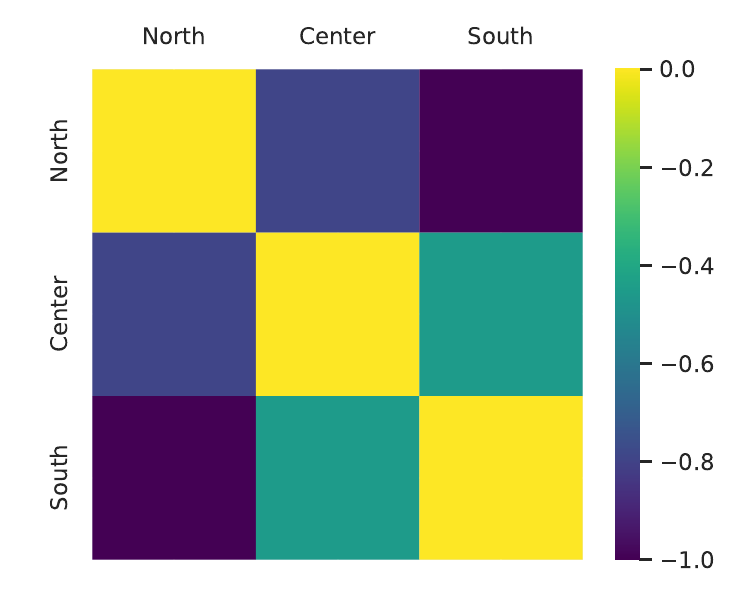}}} \\[-5pt]
            \makecell[l]{HideColors\\(\bf\textit{Ideal})}
                & \makecell{\includegraphics[clip,trim={16px 16px 16px 16px},height=48pt]{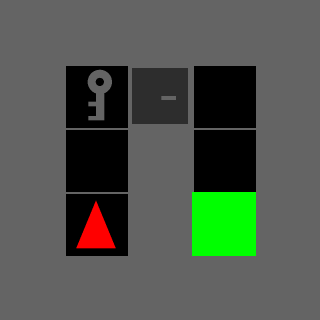}}
                & \makecell{
                    $\mathmakebox[24pt][r]{R}=$~0.000 (0.000) \\
                    $\mathmakebox[24pt][r]{\smash{\widehat{G\!R}}}=$~0.000 (0.000) \\
                    $\mathmakebox[24pt][r]{\smash{\widehat{G\!R}}{}^{\text{rec}}}=$~0.000 (0.000) \\
                    $\mathmakebox[24pt][r]{\smash{\widehat{G\!R}}{}^{\text{dec}}}=$~0.000 (0.000)}
                & \raisebox{6pt}{\makecell{\includegraphics[height=64pt]{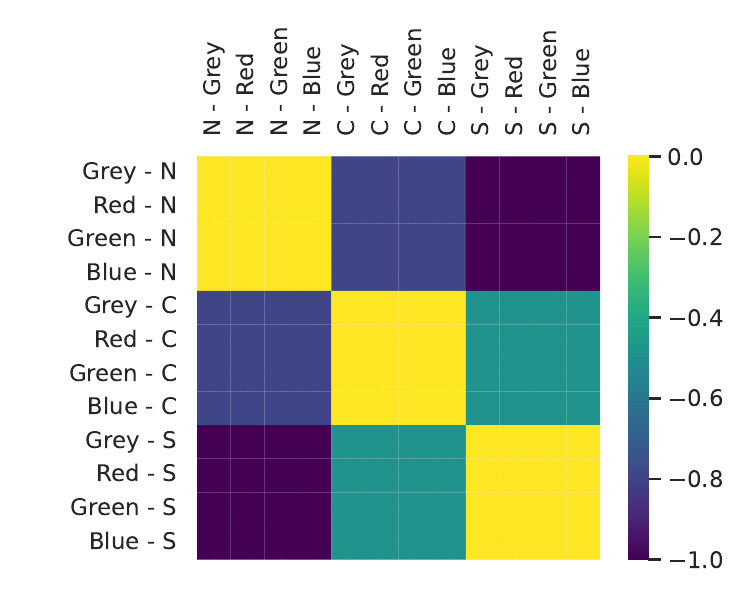}}}
                & \raisebox{1pt}{\makecell{\includegraphics[height=52pt]{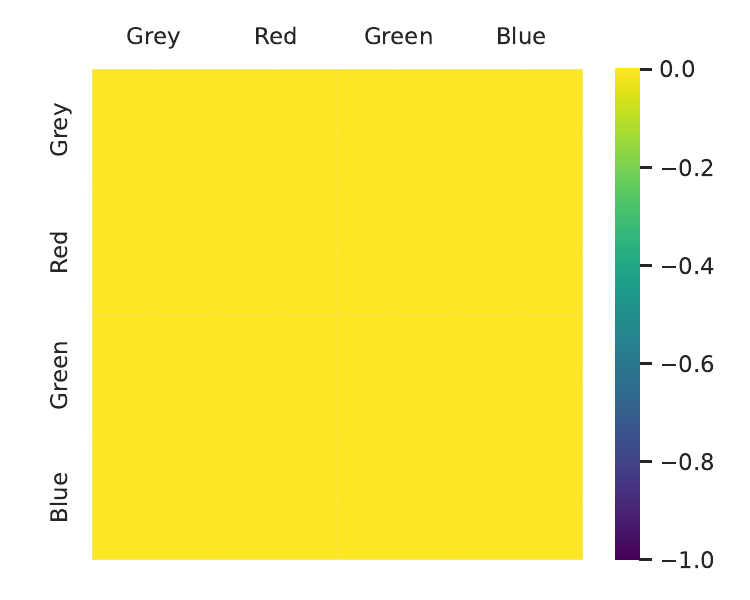}}}
                & \raisebox{1pt}{\makecell{\includegraphics[height=52pt]{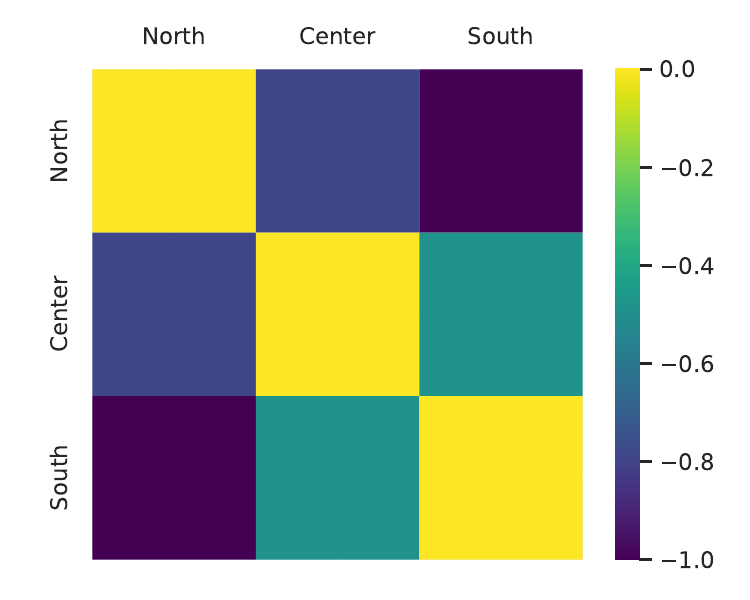}}} \\[-5pt]
            \makecell[l]{HideDoor\\(\bf\textit{Under-Specific})}
                & \makecell{\includegraphics[clip,trim={16px 16px 16px 16px},height=48pt]{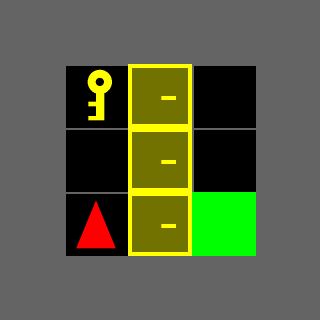}}
                & \makecell{
                    $\mathmakebox[24pt][r]{R}=$~0.070 (0.126) \\
                    $\mathmakebox[24pt][r]{\smash{\widehat{G\!R}}}=$~0.837 (0.157) \\
                    $\mathmakebox[24pt][r]{\smash{\widehat{G\!R}}{}^{\text{rec}}}=$~0.607 (0.227) \\
                    $\mathmakebox[24pt][r]{\smash{\widehat{G\!R}}{}^{\text{dec}}}=$~0.230 (0.217)}
                & \raisebox{6pt}{\makecell{\includegraphics[height=64pt]{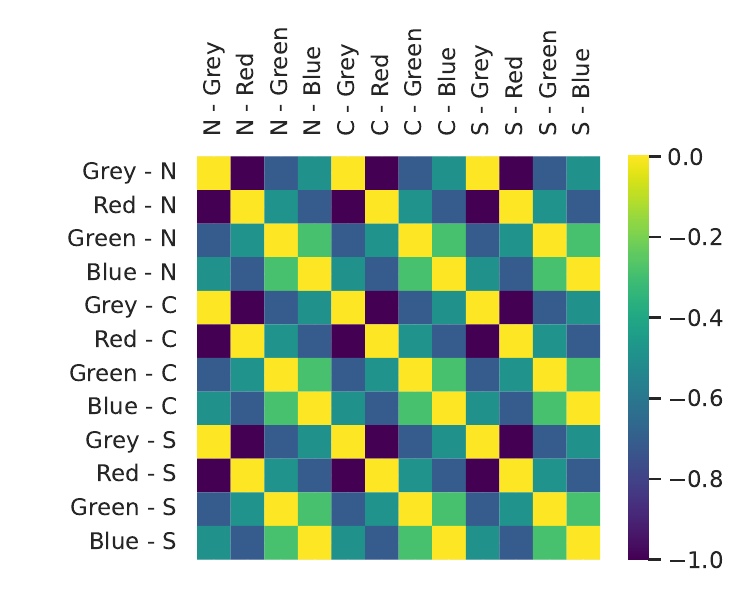}}}
                & \raisebox{1pt}{\makecell{\includegraphics[height=52pt]{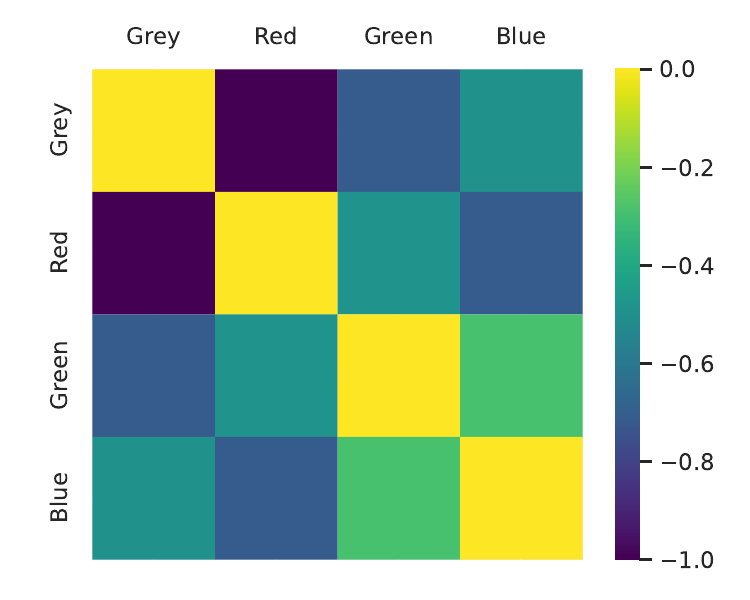}}}
                & \raisebox{1pt}{\makecell{\includegraphics[height=52pt]{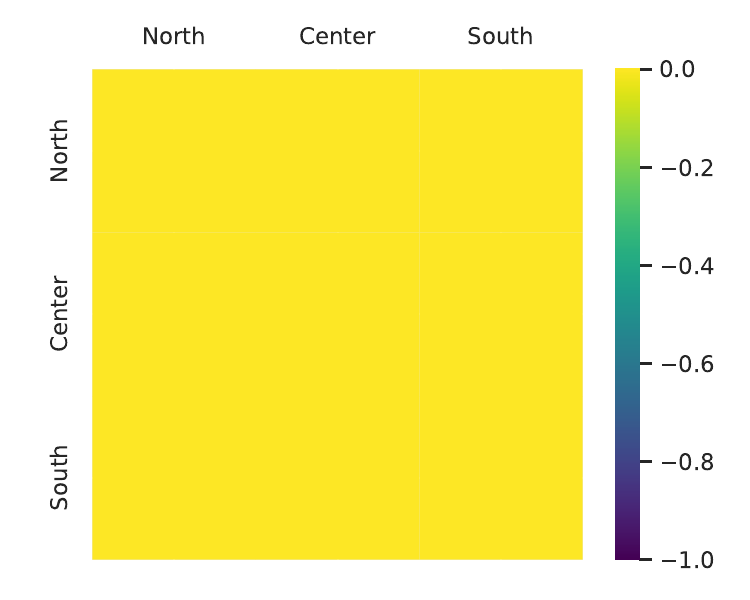}}} \\
            \bottomrule
        \end{tabular}}%
    \vspace{-\baselineskip}%
\end{table}

\looseness-1
\paragraph{Results}
Table~\ref{tab:minigrid-results} shows the regret and the generalization regret of each agent. All agents incur almost zero regret during training but only \textit{HideColors} generalizes well to the test environment.
\textit{Identity} has a higher $\smash{\widehat{G\!R}}{}^{\text{dec}}$ than $\smash{\widehat{G\!R}}{}^{\text{rec}}$ indicating that its representations are over-specific. 
This is evidenced by the dis-similarity of its representations across different maze configurations. The similarity matrices in Table~\ref{tab:minigrid-results} plot the average Euclidean distance between the representations encountered while solving one maze vs.\ another: The closer the representations of two mazes are, the more similar the agent ``thinks'' those two mazes are to each other. Representations being similar predominantly on the diagonal indicates that \textit{Identity} regard all mazes to be largely independent of each other (even more so with \textit{Gray} mazes as they are unseen during training). In contrast, \textit{HideColors} appropriately regards all mazes with the same door location as exactly the same maze regardless of the key-door color, which is ideal. 
Meanwhile, \textit{HideDoor} has a higher $\smash{\widehat{G\!R}}{}^{\text{rec}}$ than $\smash{\widehat{G\!R}}{}^{\text{dec}}$ indicating that its representations are under-specific (looking at the similarity matrices, it only differentiates between mazes with different colors).

\subsection{Examples in the Atari Game Pong}
\label{sec:contribution3b}

\begin{table}
    \centering
    \caption{All agents incur almost zero regret during training. However, \textit{NoCounter} fails to generalize to different distractions because of its over-specific representations that include irrelevant features extracted from distractions. Meanwhile, \textit{JustCounter} fails to generalize to unseen initial states because of its under-specific representations that only include the frame count as a feature, which is a spurious feature since the game is deterministic.}%
    \label{tab:pong-result}%
    \vspace{-6pt}%
    \resizebox{\linewidth-24pt}{!}{%
        \begin{tabular}{@{}lcccc@{}}
            \toprule
            \multirow{2}[3]{*}[\multirowcmidruleheight]{\bf Agent} & \multirow{2}[3]{*}[\multirowcmidruleheight]{\bf Pre-Filter} & \multirow{2}[3]{*}[\multirowcmidruleheight]{\bf\makecell{Training\\Performance}} & \multicolumn{2}{c@{}}{\bf Generalization Performance} \\
            \cmidrule(l){4-5}
            & & & \makecell{Unseen Initial States\\\small\citep[cf.][]{zhang2018dissection}} & \makecell{Different Distractions\\\small\citep[cf.][]{song2020observational}} \\
            \midrule
            Identity
                & \makecell{\includegraphics[height=42pt]{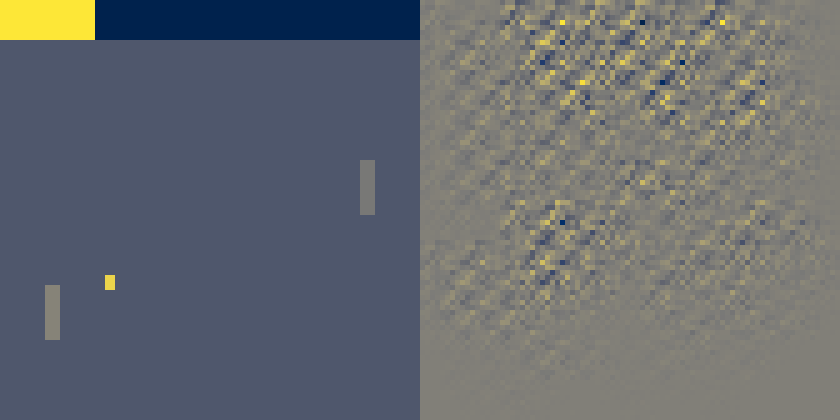}}
                & \makecell{$R=$~0.001\\\hspace{\widthof{$R=$~(0}-\widthof{(0}}(0.002)}
                & \raisebox{2pt}{\makecell{
                    $\mathmakebox[21pt][r]{\smash{\widehat{G\!R}}}=$~0.655 (0.117) \\
                    $\mathmakebox[21pt][r]{\smash{\widehat{G\!R}}^{\text{rec}}}=$~0.003 (0.003) \\
                    $\mathmakebox[21pt][r]{\smash{\widehat{G\!R}}^{\text{dec}}}=$~0.652 (0.119)}}
                & \raisebox{2pt}{\makecell{
                    $\mathmakebox[21pt][r]{\smash{\widehat{G\!R}}}=$~0.699 (0.353) \\
                    $\mathmakebox[21pt][r]{\smash{\widehat{G\!R}}^{\text{rec}}}=$~0.002 (0.002) \\
                    $\mathmakebox[21pt][r]{\smash{\widehat{G\!R}}^{\text{dec}}}=$~0.696 (0.354)}} \\
            NoCounter
                & \makecell{\includegraphics[height=42pt]{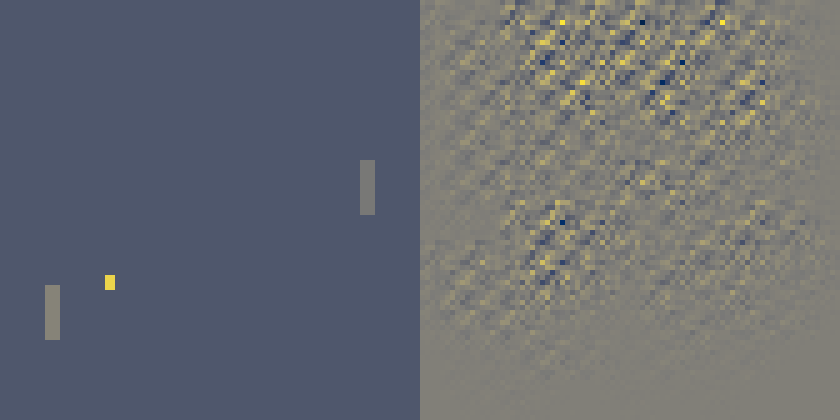}}
                & \makecell{$R=$~0.001\\\hspace{\widthof{$R=$~(0}-\widthof{(0}}(0.002)}
                & \raisebox{4pt}{\makecell{
                    \bf\textit{(Generalizes)} \\[2pt]
                    $\mathmakebox[21pt][r]{\smash{\widehat{G\!R}}}=$~0.001 (0.002) \\
                    $\mathmakebox[21pt][r]{\smash{\widehat{G\!R}}^{\text{rec}}}=$~0.000 (0.000) \\
                    $\mathmakebox[21pt][r]{\smash{\widehat{G\!R}}^{\text{dec}}}=$~0.001 (0.002)}}
                & \raisebox{4pt}{\makecell{
                    \bf\textit{(Over-Specific)} \\[2pt]
                    $\mathmakebox[21pt][r]{\smash{\widehat{G\!R}}}=$~0.642 (0.263) \\
                    $\mathmakebox[21pt][r]{\smash{\widehat{G\!R}}^{\text{rec}}}=$~0.001 (0.002) \\
                    $\mathmakebox[21pt][r]{\smash{\widehat{G\!R}}^{\text{dec}}}=$~0.641 (0.262)}} \\
            JustGameField
                & \makecell{\includegraphics[height=42pt]{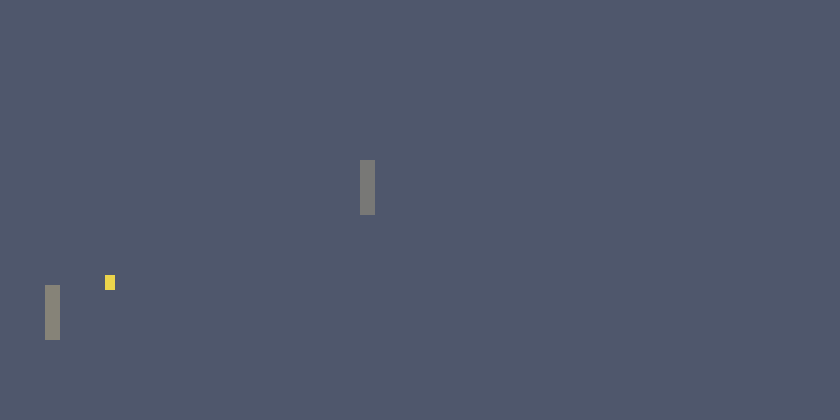}}
                & \makecell{$R=$~0.006\\\hspace{\widthof{$R=$~(0}-\widthof{(0}}(0.008)}
                & \raisebox{4pt}{\makecell{
                    \bf\textit{(Generalizes)} \\[2pt]
                    $\mathmakebox[21pt][r]{\smash{\widehat{G\!R}}}=$~0.001 (0.002) \\
                    $\mathmakebox[21pt][r]{\smash{\widehat{G\!R}}^{\text{rec}}}=$~0.001 (0.002) \\
                    $\mathmakebox[21pt][r]{\smash{\widehat{G\!R}}^{\text{dec}}}=$~0.000 (0.001)}}
                & \raisebox{4pt}{\makecell{
                    \bf\textit{(Generalizes)} \\[2pt]
                    $\mathmakebox[21pt][r]{\smash{\widehat{G\!R}}}=$~0.006 (0.008) \\
                    $\mathmakebox[21pt][r]{\smash{\widehat{G\!R}}^{\text{rec}}}=$~0.001 (0.001) \\
                    $\mathmakebox[21pt][r]{\smash{\widehat{G\!R}}^{\text{dec}}}=$~0.005 (0.008)}} \\
            JustCounter
                & \makecell{\includegraphics[height=42pt]{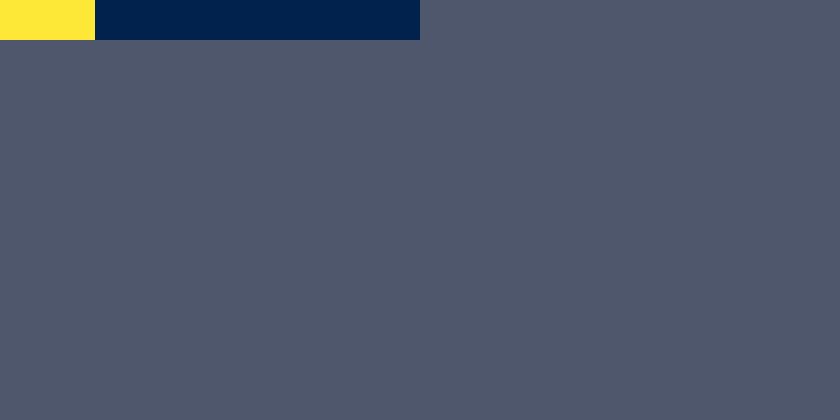}}
                & \makecell{$R=$~0.023\\\hspace{\widthof{$R=$~(0}-\widthof{(0}}(0.015)}
                & \raisebox{4pt}{\makecell{
                    \bf\textit{(Under-Specific)} \\[2pt]
                    $\mathmakebox[21pt][r]{\smash{\widehat{G\!R}}}=$~0.954 (0.006) \\
                    $\mathmakebox[21pt][r]{\smash{\widehat{G\!R}}^{\text{rec}}}=$~0.737 (0.152) \\
                    $\mathmakebox[21pt][r]{\smash{\widehat{G\!R}}^{\text{dec}}}=$~0.217 (0.156)}}
                & \raisebox{4pt}{\makecell{
                    \bf\textit{(Generalizes)} \\[2pt]
                    $\mathmakebox[21pt][r]{\smash{\widehat{G\!R}}}=$~0.023 (0.015) \\
                    $\mathmakebox[21pt][r]{\smash{\widehat{G\!R}}^{\text{rec}}}=$~0.002 (0.003) \\
                    $\mathmakebox[21pt][r]{\smash{\widehat{G\!R}}^{\text{dec}}}=$~0.021 (0.016)}} \\
            \bottomrule
        \end{tabular}}%
\end{table}

\paragraph{Setup}
We consider a deterministic version of Pong and modify the game so that it is reset as soon as the first point is scored. Similar to \textit{Minigrid}, creating agents with over/under-specific representations requires us to introduce irrelevant and spurious features to the game. As the spurious features, we add a frame counter on the top of the game field in the shape of a progress bar (see \textit{Identity} in Table~\ref{tab:pong-result}). Since the game is completely deterministic, it should be possible to play the game successfully based only on this counter, making it a spurious feature correlated with the optimal action trajectories. However, such a strategy should not generalize to a new initial state \citep[as in the framework of][]{zhang2018dissection}. In our experiments, we will vary the initial state by taking a random number of null actions at the start of the game without actually increasing the frame counter.

As the irrelevant features, we generate synthetic images and concatenate them with the original images from the game. We refer to these synthetic images as distractions. They are generated via deterministic functions of the game state and hence contain exactly the same information as the original game screen. Whenever the game is reset, we pick one of two possible generating functions at random, making distractions analogous to decorative elements in other games that change from one level to another \citep[as in the framework of][]{song2020observational}. Agents that focus on distractions rather than the proper game field should not generalize to a different set of distractions. In our experiments, we will vary the set of two possible generating functions from training to test time.

\looseness-1
\paragraph{Agents}
Similar to before, we construct different agents via different pre-filters (see the corresponding column in Table~\ref{tab:pong-result}): (i) \textit{Identity} makes no modifications to the input image. (ii) \textit{NoCounter} hides the frame counter keeping the distractions. This should leave the agent susceptible to learning over-specific representations that include irrelevant features extracted from distractions. (iii) \textit{JustCounter} hides everything but the game counter. This should leave the agent susceptible to learning under-specific representations that only include a spurious feature, the frame count. (iv)~\mbox{\textit{JustGameField}} hides both the frame counter (spurious) and any distractions (irrelevant), which is ideal.

\paragraph{Results}
Table~\ref{tab:pong-result} shows the regret and the generalization regret, both for generalization to unseen initial states and different distractions. All agents incur almost zero regret during training. However, \textit{NoCounter} does not generalize to different distractions. This is because the representations learned by \textit{NoCounter} are over-specific (high $\smash{\widehat{G\!R}}{}^{\text{dec}}$ over $\smash{\widehat{G\!R}}{}^{\text{rec}}$). Meanwhile, \textit{JustCounter} does not generalize to unseen initial states. This is because the representations learned by \textit{JustCounter} are under-specific (high $\smash{\widehat{G\!R}}{}^{\text{rec}}$ over $\smash{\widehat{G\!R}}{}^{\text{dec}}$).

\vspace{-5pt}
\subsection{Use Case: Model Selection}
\label{sec:contribution3c}
\vspace{-5pt}

\begin{wrapfigure}[13]{r}{.52\linewidth}
    \vspace{-\baselineskip}%
    \centering
    \includegraphics[width=.5\linewidth,clip,trim={9pt 12pt 9pt 12pt}]{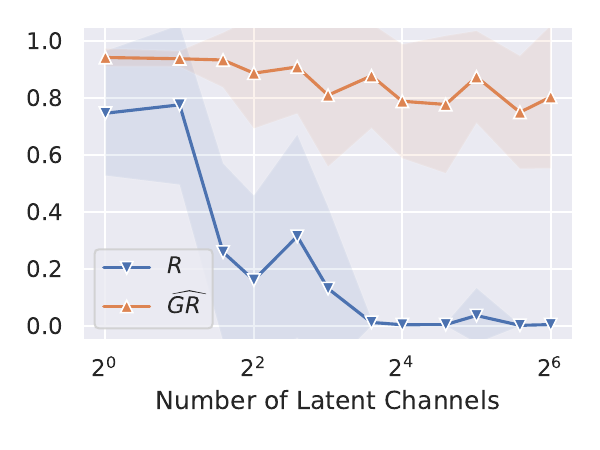}%
    \includegraphics[width=.5\linewidth,clip,trim={9pt 12pt 9pt 12pt}]{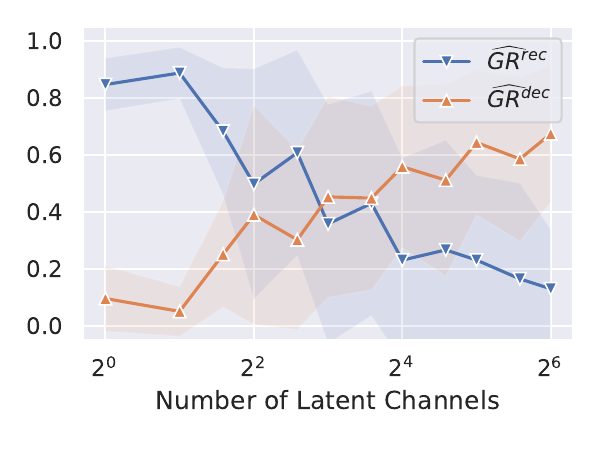}%
    \caption{\textit{Regret and Generalization Regrets of Identity in Maze Environments.} Reducing the number of latent channels prevents over-specific representations and lowers $\widehat{G\!R}{}^{\text{dec}}$. However, this also results in the recognition policy overfitting to the key-door color, ultimately leading to no improvements in terms of the overall generalization regret.}%
    \label{fig:usecase}
\end{wrapfigure}

Analyzing the recognition and decision generalization regrets of RL models can help make informed decisions about how to improve generalization. If generalization performance is poor due to over-specific representations, we may  switch to an algorithm that is capable of learning invariances within the observation space \citep[e.g.][]{sonar2021invariant}. Conversely, if our representations are under-specific, we may introduce additional loss terms to encourage the model to extract task-critical features.

As a concrete example, we consider one of the simplest adjustments to make to the model architecture: changing the representation dimensionality. We consider \textit{Identity} from the examples in maze environments, whose recognition policy outputs a latent representation with $2^6=64$ channels. Observing that \textit{Identity} has a high $\widehat{G\!R}{}^{\text{dec}}$ relative to $\widehat{G\!R}{}^{\text{rec}}$, we conclude that generalization fails due to over-specific representations. To discourage the model from extracting irrelevant features, we decide to reduce the number of latent channels. We can see in Figure~\ref{fig:usecase} that our intervention is successful in lowering $\widehat{G\!R}{}^{\text{dec}}$, but the model then ends up overfitting to spurious features like the key-door color instead, resulting in high $\widehat{G\!R}{}^{\text{rec}}$ and no overall improvement in terms of the generalization regret. Now that we know our setting suffers both from irrelevant and spurious features, we can start investigating alternative ways to address either one of these problematic features, adopting an appropriate number of latent channels to the issue we target. For instance, we can also see in Figure~\ref{fig:usecase} that the increase in $\widehat{G\!R}{}^{\text{rec}}$ is rather marginal until the number of latent channels is reduced to $2^4=16$. We can set this as our new representation size, knowing it is sufficiently large to capture all task-critical information, and decide to address the issue of irrelevant features first.

\vspace{-5pt}
\section{Conclusion}
\vspace{-5pt}

In this paper, we introduced the notions of recognition and decision regrets, which we used to analyze the generalization performance of RL agents. Through illustrative examples, we demonstrated how these regret metrics can characterize under-specific and over-specific representations, offering a more nuanced perspective on observational overfitting in image-based RL. We highlighted how such characterization could be helpful in systematically improving the generalization performance of RL models. Future research should focus on identifying specific strategies that one can follow when an RL model fails to generalize due to over-specific vs.\ under-specific representations. Our code is publicly available at \url{https://github.com/alihanhyk/disentangling-regret}.


\appendix

\bibliography{references}

\begin{thebibliography}{30}
\providecommand{\natexlab}[1]{#1}
\providecommand{\url}[1]{\texttt{#1}}
\expandafter\ifx\csname urlstyle\endcsname\relax
  \providecommand{\doi}[1]{DOI: #1}\else
  \providecommand{\doi}{DOI: \begingroup \urlstyle{rm}\Url}\fi

\bibitem[Agarwal et~al.(2021)Agarwal, Machado, Castro, and
  Bellemare]{agarwal2021contrastive}
Rishabh Agarwal, Marlos~C Machado, Pablo~Samuel Castro, and Marc~G Bellemare.
\newblock Contrastive behavioral similarity embeddings for generalization in
  reinforcement learning.
\newblock In \emph{International Conference on Learning Representations}, 2021.

\bibitem[Bellemare et~al.(2019)Bellemare, Dabney, Dadashi, Ali~Taiga, Castro,
  Le~Roux, Schuurmans, Lattimore, and Lyle]{bellemare2019geometric}
Marc Bellemare, Will Dabney, Robert Dadashi, Adrien Ali~Taiga, Pablo~Samuel
  Castro, Nicolas Le~Roux, Dale Schuurmans, Tor Lattimore, and Clare Lyle.
\newblock A geometric perspective on optimal representations for reinforcement
  learning.
\newblock In \emph{Conference on Neural Information Processing Systems}, 2019.

\bibitem[Bellemare et~al.(2013)Bellemare, Naddaf, Veness, and
  Bowling]{bellemare2013arcade}
Marc~G Bellemare, Yavar Naddaf, Joel Veness, and Michael Bowling.
\newblock The arcade learning environment: An evaluation platform for general
  agents.
\newblock \emph{Journal of Artificial Intelligence Research}, 47:\penalty0
  253--279, 2013.

\bibitem[Chevalier-Boisvert et~al.(2023)Chevalier-Boisvert, Dai, Towers,
  de~Lazcano, Willems, Lahlou, Pal, Castro, and Terry]{maxime2023minigrid}
Maxime Chevalier-Boisvert, Bolun Dai, Mark Towers, Rodrigo de~Lazcano, Lucas
  Willems, Salem Lahlou, Suman Pal, Pablo~Samuel Castro, and Jordan Terry.
\newblock {Minigrid} \& {Miniworld}: Modular \& customizable reinforcement
  learning environments for goal-oriented tasks.
\newblock \emph{CoRR}, abs/2306.13831, 2023.

\bibitem[Cobbe et~al.(2019)Cobbe, Klimov, Hesse, Kim, and
  Schulman]{cobbe2019quantifying}
Karl Cobbe, Oleg Klimov, Chris Hesse, Taehoon Kim, and John Schulman.
\newblock Quantifying generalization in reinforcement learning.
\newblock In \emph{International Conference on Machine Learning}, pp.\
  1282--1289, 2019.

\bibitem[Dabney et~al.(2021)Dabney, Barreto, Rowland, Dadashi, Quan, Bellemare,
  and Silver]{dabney2021value}
Will Dabney, Andr{\'e} Barreto, Mark Rowland, Robert Dadashi, John Quan, Marc~G
  Bellemare, and David Silver.
\newblock The value-improvement path: Towards better representations for
  reinforcement learning.
\newblock In \emph{Proceedings of the AAAI Conference on Artificial
  Intelligence}, 2021.

\bibitem[Eysenbach et~al.(2022)Eysenbach, Zhang, Levine, and
  Salakhutdinov]{eysenbach2022contrastive}
Benjamin Eysenbach, Tianjun Zhang, Sergey Levine, and Russ~R Salakhutdinov.
\newblock Contrastive learning as goal-conditioned reinforcement learning.
\newblock In \emph{Conference on Neural Information Processing Systems}, 2022.

\bibitem[Ghosh \& Bellemare(2020)Ghosh and Bellemare]{ghosh2020representations}
Dibya Ghosh and Marc~G Bellemare.
\newblock Representations for stable off-policy reinforcement learning.
\newblock In \emph{International Conference on Machine Learning}, 2020.

\bibitem[Jiang et~al.(2015)Jiang, Kulesza, Singh, and
  Lewis]{jiang2015dependence}
Nan Jiang, Alex Kulesza, Satinder Singh, and Richard Lewis.
\newblock The dependence of effective planning horizon on model accuracy.
\newblock In \emph{Proceedings of the 2015 international conference on
  autonomous agents and multiagent systems}, pp.\  1181--1189, 2015.

\bibitem[Le~Lan et~al.(2022)Le~Lan, Tu, Oberman, Agarwal, and
  Bellemare]{le2022generalization}
Charline Le~Lan, Stephen Tu, Adam Oberman, Rishabh Agarwal, and Marc~G
  Bellemare.
\newblock On the generalization of representations in reinforcement learning.
\newblock In \emph{International Conference on Artificial Intelligence and
  Statistics}, 2022.

\bibitem[Machado et~al.(2018)Machado, Bellemare, Talvitie, Veness, Hausknecht,
  and Bowling]{machado2018revisiting}
Marlos~C Machado, Marc~G Bellemare, Erik Talvitie, Joel Veness, Matthew
  Hausknecht, and Michael Bowling.
\newblock Revisiting the arcade learning environment: Evaluation protocols and
  open problems for general agents.
\newblock \emph{Journal of Artificial Intelligence Research}, 61:\penalty0
  523--562, 2018.

\bibitem[Mnih et~al.(2015)Mnih, Kavukcuoglu, Silver, Rusu, Veness, Bellemare,
  Graves, Riedmiller, Fidjeland, Ostrovski, et~al.]{mnih2015human}
Volodymyr Mnih, Koray Kavukcuoglu, David Silver, Andrei~A Rusu, Joel Veness,
  Marc~G Bellemare, Alex Graves, Martin Riedmiller, Andreas~K Fidjeland, Georg
  Ostrovski, et~al.
\newblock Human-level control through deep reinforcement learning.
\newblock \emph{Nature}, 518\penalty0 (7540):\penalty0 529--533, 2015.

\bibitem[Mnih et~al.(2016)Mnih, Badia, Mirza, Graves, Lillicrap, Harley,
  Silver, and Kavukcuoglu]{mnih2016asynchronous}
Volodymyr Mnih, Adria~Puigdomenech Badia, Mehdi Mirza, Alex Graves, Timothy
  Lillicrap, Tim Harley, David Silver, and Koray Kavukcuoglu.
\newblock Asynchronous methods for deep reinforcement learning.
\newblock In \emph{International Conference on Machine Learning}, pp.\
  1928--1937, 2016.

\bibitem[Nichol et~al.(2018)Nichol, Pfau, Hesse, Klimov, and
  Schulman]{nichol2018gotta}
Alex Nichol, Vicki Pfau, Christopher Hesse, Oleg Klimov, and John Schulman.
\newblock Gotta learn fast: A new benchmark for generalization in rl.
\newblock \emph{arXiv preprint arXiv:1804.03720}, 2018.

\bibitem[Packer et~al.(2018)Packer, Gao, Kos, Kr{\"a}henb{\"u}hl, Koltun, and
  Song]{packer2018assessing}
Charles Packer, Katelyn Gao, Jernej Kos, Philipp Kr{\"a}henb{\"u}hl, Vladlen
  Koltun, and Dawn Song.
\newblock Assessing generalization in deep reinforcement learning.
\newblock \emph{arXiv preprint arXiv:1810.12282}, 2018.

\bibitem[Pinto et~al.(2017)Pinto, Davidson, Sukthankar, and
  Gupta]{pinto2017robust}
Lerrel Pinto, James Davidson, Rahul Sukthankar, and Abhinav Gupta.
\newblock Robust adversarial reinforcement learning.
\newblock In \emph{International Conference on Machine Learning}, pp.\
  2817--2826, 2017.

\bibitem[Raffin et~al.(2021)Raffin, Hill, Gleave, Kanervisto, Ernestus, and
  Dormann]{raffin2021stable}
Antonin Raffin, Ashley Hill, Adam Gleave, Anssi Kanervisto, Maximilian
  Ernestus, and Noah Dormann.
\newblock Stable-baselines3: Reliable reinforcement learning implementations.
\newblock \emph{Journal of Machine Learning Research}, 22\penalty0
  (268):\penalty0 1--8, 2021.

\bibitem[Raileanu \& Fergus(2021)Raileanu and Fergus]{raileanu2021decoupling}
Roberta Raileanu and Rob Fergus.
\newblock Decoupling value and policy for generalization in reinforcement
  learning.
\newblock In \emph{International Conference on Machine Learning}, 2021.

\bibitem[Rajeswaran et~al.(2017)Rajeswaran, Lowrey, Todorov, and
  Kakade]{rajeswaran2017towards}
Aravind Rajeswaran, Kendall Lowrey, Emanuel~V Todorov, and Sham~M Kakade.
\newblock Towards generalization and simplicity in continuous control.
\newblock \emph{Neural Information Processing Systems}, 30, 2017.

\bibitem[Schulman et~al.(2017)Schulman, Wolski, Dhariwal, Radford, and
  Klimov]{schulman2017proximal}
John Schulman, Filip Wolski, Prafulla Dhariwal, Alec Radford, and Oleg Klimov.
\newblock Proximal policy optimization algorithms.
\newblock \emph{arXiv preprint arXiv:1707.06347}, 2017.

\bibitem[Serrano-Muñoz et~al.(2023)Serrano-Muñoz, Chrysostomou, Bøgh, and
  Arana-Arexolaleiba]{serrano2023skrl}
Antonio Serrano-Muñoz, Dimitrios Chrysostomou, Simon Bøgh, and Nestor
  Arana-Arexolaleiba.
\newblock skrl: Modular and flexible library for reinforcement learning.
\newblock \emph{Journal of Machine Learning Research}, 24, 2023.

\bibitem[Sonar et~al.(2021)Sonar, Pacelli, and Majumdar]{sonar2021invariant}
Anoopkumar Sonar, Vincent Pacelli, and Anirudha Majumdar.
\newblock Invariant policy optimization: Towards stronger generalization in
  reinforcement learning.
\newblock In \emph{Learning for Dynamics and Control}, 2021.

\bibitem[Song et~al.(2020)Song, Jiang, Tu, Du, and
  Neyshabur]{song2020observational}
Xingyou Song, Yiding Jiang, Stephen Tu, Yilun Du, and Behnam Neyshabur.
\newblock Observational overfitting in reinforcement learning.
\newblock In \emph{International Conference on Learning Representations}, 2020.

\bibitem[Stooke et~al.(2021)Stooke, Lee, Abbeel, and
  Laskin]{stooke2021decoupling}
Adam Stooke, Kimin Lee, Pieter Abbeel, and Michael Laskin.
\newblock Decoupling representation learning from reinforcement learning.
\newblock In \emph{International Conference on Machine Learning}, 2021.

\bibitem[Taiga et~al.(2023)Taiga, Agarwal, Farebrother, Courville, and
  Bellemare]{taiga2023investigating}
Adrien~Ali Taiga, Rishabh Agarwal, Jesse Farebrother, Aaron Courville, and
  Marc~G Bellemare.
\newblock Investigating multi-task pretraining and generalization in
  reinforcement learning.
\newblock In \emph{International Conference on Learning Representations}, 2023.

\bibitem[Towers et~al.(2023)Towers, Terry, Kwiatkowski, Balis, Cola, Deleu,
  Goulão, Kallinteris, KG, Krimmel, Perez-Vicente, Pierré, Schulhoff, Tai,
  Shen, and Younis]{towers_gymnasium_2023}
Mark Towers, Jordan~K. Terry, Ariel Kwiatkowski, John~U. Balis, Gianluca~de
  Cola, Tristan Deleu, Manuel Goulão, Andreas Kallinteris, Arjun KG, Markus
  Krimmel, Rodrigo Perez-Vicente, Andrea Pierré, Sander Schulhoff, Jun~Jet
  Tai, Andrew Tan~Jin Shen, and Omar~G. Younis.
\newblock Gymnasium, March 2023.
\newblock URL \url{https://zenodo.org/record/8127025}.

\bibitem[Wang et~al.(2024)Wang, Miahi, White, Machado, Abbas, Kumaraswamy, Liu,
  and White]{wang2024investigating}
Han Wang, Erfan Miahi, Martha White, Marlos~C Machado, Zaheer Abbas, Raksha
  Kumaraswamy, Vincent Liu, and Adam White.
\newblock Investigating the properties of neural network representations in
  reinforcement learning.
\newblock \emph{Artificial Intelligence}, 330:\penalty0 104100, 2024.

\bibitem[Zhang et~al.(2018{\natexlab{a}})Zhang, Ballas, and
  Pineau]{zhang2018dissection}
Amy Zhang, Nicolas Ballas, and Joelle Pineau.
\newblock A dissection of overfitting and generalization in continuous
  reinforcement learning.
\newblock \emph{arXiv preprint arXiv:1806.07937}, 2018{\natexlab{a}}.

\bibitem[Zhang et~al.(2018{\natexlab{b}})Zhang, Vinyals, Munos, and
  Bengio]{zhang2018study}
Chiyuan Zhang, Oriol Vinyals, Remi Munos, and Samy Bengio.
\newblock A study on overfitting in deep reinforcement learning.
\newblock \emph{arXiv preprint arXiv:1804.06893}, 2018{\natexlab{b}}.

\bibitem[Zhang et~al.(2022)Zhang, Song, Uehara, Wang, Agarwal, and
  Sun]{zhang2022efficient}
Xuezhou Zhang, Yuda Song, Masatoshi Uehara, Mengdi Wang, Alekh Agarwal, and Wen
  Sun.
\newblock Efficient reinforcement learning in block {MDP}s: A model-free
  representation learning approach.
\newblock In \emph{International Conference on Machine Learning}, 2022.

\end{thebibliography}
\bibliographystyle{rlj}

\clearpage
\allowdisplaybreaks

\section{A Worked Example for Generalization}
\label{sec:appendix-workedexample}

\begin{figure}[h]
    \centering%
    \resizebox{.5\linewidth}{!}{%
        \begin{tikzpicture}[auto, >=latex,font=\small, pin distance=10mm]
        
        \node[on grid] (o0) {\includegraphics[width=20mm]{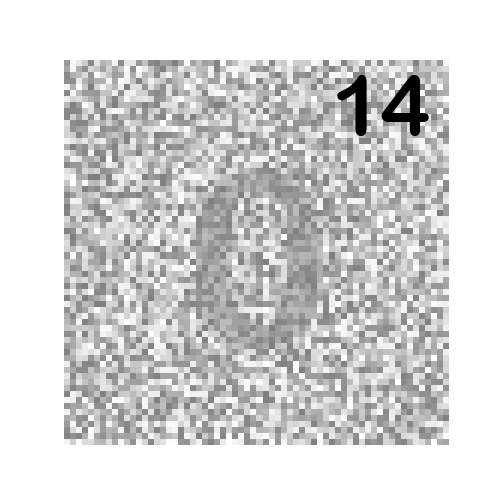}};

        \node[on grid,right=20mm of o0] (o1) {\includegraphics[width=20mm]{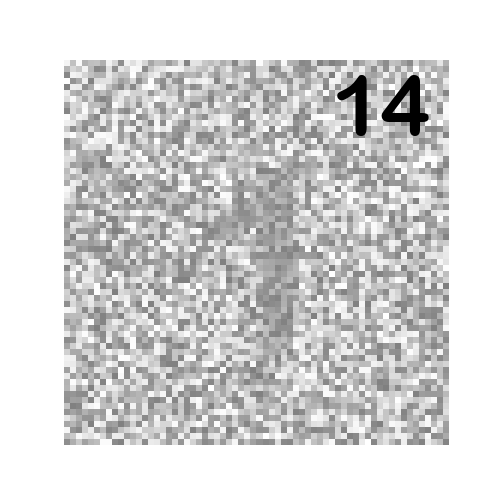}};

        \node[on grid,right=20mm of o1] (o2) {\includegraphics[width=20mm]{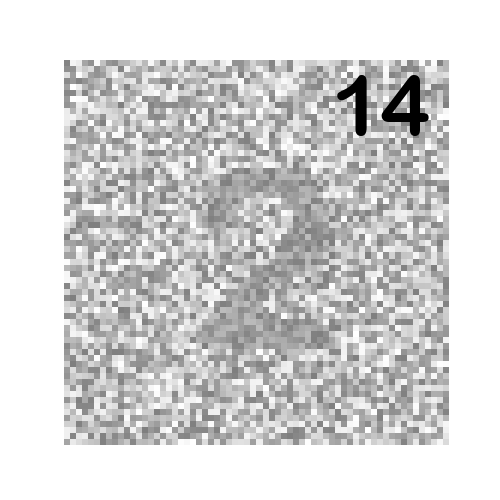}};
    \end{tikzpicture}}%
    \vspace{-12pt}%
    \caption{Modified observation space $X^{\text{gray}}$ (at time step $t=14$).}%
    \label{fig:worked_example_test_images}
\end{figure}

We now show how these definitions apply to our first worked example. The training environment $\varepsilon^{\text{train}}$ is exactly same as the original environment depicted in Figure~\ref{fig:worked_example}. The test environment $\varepsilon^{\text{test}}$ differs in two ways. First, we modify the observation space $X$ to a new space $X^{\text{gray}}$ where the figures in the images are converted from red/green/blue to grayscale, as illustrated in Figure~\ref{fig:worked_example_test_images}. Second, we start the environment from state $0$ instead of state $1$. The optimal decision policy $\pi^*$ (i.e.\ the best action in each state) is the same in $\varepsilon^{\text{test}}$ as in $\varepsilon^{\text{train}}$, but due to the change in initial state the value is now $V_{\varepsilon^{\text{test}}}^*=\gamma+\gamma^3+\ldots,$ which equals $4.737$ when $\gamma=0.9$. 
Note that a recognition policy $\rho_0$ can achieve optimal performance in the training environment, i.e.\ $\max_{\pi} V_{\varepsilon^{\text{train}}}(\rho_0,\pi)=V_{\varepsilon^{\text{train}}}^*$, whenever $s_t=i$ implies $\pi'_0(x_t)=i$ for $x_t\in X$. However, different recognition policies satisfying that condition may have differing behavior in the test environment (i.e.\ for $x_t\in X^{\text{gray}}$).   

\paragraph{Example of Under-Specific Recognition Policy}
Suppose that $\rho'_0(\cdot)$ makes its decision based on whether there are any red, green, or blue pixels in the image, and makes a random decision otherwise. In other words, for the test environment the recognition policy does not given any useful information about the state (since the observation is grayscale). Intuitively, this is why it is deemed under-specified but we now show that it matches the definition of the term that we gave previously. 

If we apply the decision policy that was optimal for $\varepsilon^{\text{test}}$ then effectively in all states we take action $0$ with probability $1/3$ and action $1$ with probability $2/3$ (since the representation state $z_t$ is independent from $s_t$). 
The value in the environment for these policies is given by $v_1$ in the equations $v_0=\gamma v_0(1/3)+\gamma v_1(2/3)$, $v_1=\gamma v_0(1/3)+\gamma v_2(2/3)$, $v_2 = 1+\gamma v_0(2/3)+\gamma v_1(1/3)$. When $\gamma=0.9$, the solution to these equations is $(v_0,v_1,v_2)=(2.022,2.360,2.921)$ hence the generalization regret:
\begin{align}
    {G\!R} = {\max}{}_{\rho,\pi}V_{\varepsilon^{\text{test}}}(\rho,\pi) - V_{\varepsilon^{\text{test}}}(\argmax{}_{\pi}V_{\varepsilon^{\text{train}}}(\rho_0,\pi))=4.737-2.360=2.377
\end{align}
Note that for this simple case, since we can calculate optimal policy in $\varepsilon^{\text{test}}$ exactly, there is no difference between $\smash{G\!R{}}$ and the empirical generalization regret $\smash{\widehat{G\!R}}$ and so we will only focus on the former in this section. 

Since the generalization regret is high, it is simple to obtain zero regret in $\varepsilon^{\text{train}}$, and the only difference between $\varepsilon^{\text{test}}$ and $\varepsilon^{\text{train}}$ is in the observation dynamics, this situation matches our definition of observational overfitting.
We now decompose the generalization regret $\smash{G\!R{}}$ into its components $\smash{G\!R{}^{\text{rec}}}$ and $\smash{G\!R{}^{\text{dec}}}$. When the recognition policy $\rho_0$ is used, the {\em optimal} decision policy for the test environment $\argmax{}_{\pi}V_{\varepsilon^{\text{test}}}(\rho_0,\pi)$ is simply the optimal decision policy that is not aware of the state. It is not hard to verify that this decision policy always takes action $1$ and so $\max{}_{\pi}V_{\varepsilon^{\text{test}}}(\rho_0,\pi)=\gamma+\gamma^4+\ldots,$ which equals $2.989$ when $\gamma=0.9$. Hence,
\begin{align}
    G\!R{}^{\text{rec}} &= V_{\varepsilon^{\text{test}}}(\argmax{}_{\rho,\pi}V_{\varepsilon^{\text{test}}}(\rho,\pi)) - V_{\varepsilon^{\text{test}}}(\rho_0,\argmax{}_{\pi}V_{\varepsilon^{\text{test}}}(\rho_0,\pi)) \nonumber \\
    &= 4.737-2.989 \nonumber \\
    &= 1.748 \\
    G\!R{}^{\text{dec}} &= V_{\varepsilon^{\text{test}}}(\rho_0,\argmax{}_{\pi}V_{\varepsilon^{\text{test}}}(\rho_0,\pi)) - V_{\varepsilon^{\text{test}}}(\rho_0,\argmax{}_{\pi}V_{\varepsilon^{\text{train}}}(\rho_0,\pi)) \nonumber \\
    &= 2.989-2.360 \nonumber \\
    &= 0.629
\end{align}
Since the majority of the regret comes from $G\!R{}^{\text{rec}}$, this matches our definition of the recognition policy $\rho_0$
being {\em under-specified}. This is intuitive since by focusing on the colors, the recognition policy $\rho_0$  does not recognize the salient feature of the observation state in $\varepsilon^{\text{test}}$ (i.e.\ whether the number in the image is $0$, $1$ or $2$). 

\paragraph{Example of Over-Specific Recognition Policy}
We now consider a different recognition policy $\rho_1$ that focuses on the timestamp in the top-right corner of each image in Figure~\ref{fig:worked_example}. In particular the representation state is defined by $\rho'_1(x_t)=0$ if $t=0$, otherwise $\rho'_1(x_t)=1$ if $t$ is odd and $\rho'_1(x_t)=2$ if $t$ is even. This allows us to obtain the optimal reward in $\varepsilon^{\text{train}}$ since if we apply the optimal decision policy (that takes action $1$ in states $0,1$ and action $0$ in state $1$), then we will follow the optimal trajectory and  $s_t=\rho'_1(x_t)$ at all times $t$. Note that in this optimal trajectory the action sequence is $1,1,0,1,0,1,\ldots$

However, if we move to the test environment $\varepsilon^{\text{test}}$, and apply the same recognition and decision policies, we will follow the exact same action sequence $1,1,0,1,0,1,\ldots$ since the recognition policy is only looking at the timestamp. However, due to the change in initial state, this is not optimal in $\varepsilon^{\text{test}}$. In particular, the state sequence becomes $1,2,0,0,1,0,1,0,\ldots$, i.e.\ we remain in a loop between states $0,1$ rather than the states $1,2$ where we want to be to receive the reward. The value of this sequence is $\gamma=0.9$. 
Hence the generalization regret:
\begin{align}
    {G\!R} &= {\max}{}_{\rho,\pi}V_{\varepsilon^{\text{test}}}(\rho,\pi) - V_{\varepsilon^{\text{test}}}(\argmax{}_{\pi}V_{\varepsilon^{\text{train}}}(\rho_0,\pi))=4.737-0.900=3.837
\end{align}

However, there is clearly another decision policy that still obtains the optimal reward in $\varepsilon^{\text{test}}$ even with the recognition policy $\rho_1$. We just need to take action $1$ when $\rho'_1(x_t)=0$, action $0$ when $\rho'_1(x_t)=1$, and action $1$ when $\rho'_1(x_t)=2$. (It can be checked that this leads to the optimal action sequence $1,0,1,0,\ldots$) Hence, 
\begin{align}
    G\!R{}^{\text{rec}} &= V_{\varepsilon^{\text{test}}}(\argmax{}_{\rho,\pi}V_{\varepsilon^{\text{test}}}(\rho,\pi)) - V_{\varepsilon^{\text{test}}}(\rho_1,\argmax{}_{\pi}V_{\varepsilon^{\text{test}}}(\rho_1,\pi)) \nonumber \\
    &= 4.737-4.737 \nonumber \\
    &= 0 \\
    G\!R{}^{\text{dec}} &= V_{\varepsilon^{\text{test}}}(\rho_1,\argmax{}_{\pi}V_{\varepsilon^{\text{test}}}(\rho_1,\pi)) - V_{\varepsilon^{\text{test}}}(\rho_1,\argmax{}_{\pi}V_{\varepsilon^{\text{train}}}(\rho_1,\pi)) \nonumber \\
    &= 4.737-0.900 \nonumber \\
    &= 3.837 
\end{align}
In this case all of the generalization regret is concentrated on the decision regret, and so this matches our definition of the recognition policy $\rho_1$
being {\em over-specified}. This is intuitive since by focusing on the timestamp, the recognition policy $\rho_1$ is basing its decision on a feature that is specific to a single image, rather than the number in the image which is common to a whole image class. 


\section{Extended Results for Maze Environments}
\label{sec:appendix-extendedresults}

In addition to three agent in Section~\ref{sec:contribution3a}, we consider two more agent with more extreme properties: (iv) \textit{Blind} combines both \textit{HideDoor} and \textit{HideColors}, resulting in extremely under-specific representations. (v) \textit{HideDoor} encodes each object-color combination as its own distinct token in a one-hot encoding scheme. Since object types, states, and colors are no longer encoded in separate channels, \textit{OneHot} makes it much harder to learn recognition policies that ignore color, resulting in extremely over-specific representations.

An extended version of Table~\ref{tab:minigrid-results} with these two additional agents is given in Table~\ref{tab:minigrid-resultsextended}. Notably, the diagonal in the similarity matrices of \textit{Identity} is even more pronounced for \textit{OneHot}. This indicates that \textit{OneHot} considers each maze in isolation without relating any features between different configurations, even more so than \textit{Identity}. Meanwhile, \textit{Blind} incurs a high regret even during training as it receives practically no information to base its actions on. 

\begin{table}
    \centering%
    \caption{Extended results for \textit{Minigrid} in Section~\ref{sec:contribution3b}.}%
    \label{tab:minigrid-resultsextended}%
    \resizebox{\linewidth-24pt}{!}{%
        \begin{tabular}{@{}lccc@{}c@{}c@{}}
            \toprule
            \multirow{2}{*}[\multirowcmidruleheight]{\bf Agent} & \multirow{2}{*}[\multirowcmidruleheight]{\bf Pre-Filter} & \multirow{2}{*}[\multirowcmidruleheight]{\bf Performance} & \multicolumn{3}{c@{\hspace{-10pt}}}{\bf Representation Similarity} \\
            \cmidrule(l){4-6}
            & & & All & By Color & By Door Location \\
            \midrule
            \makecell[l]{OneHot\\(\bf\textit{Extremely}\\\hspace{3pt}\bf\textit{Over-Specific})}
                & \textcolor{black!20}{[N/A]}
                & \makecell{
                    $\mathmakebox[24pt][r]{R}=$~0.002 (0.003) \\
                    $\mathmakebox[24pt][r]{\smash{\widehat{G\!R}}}=$~0.968 (0.059) \\
                    $\mathmakebox[24pt][r]{\smash{\widehat{G\!R}}^{\text{rec}}}=$~0.536 (0.342) \\
                    $\mathmakebox[24pt][r]{\smash{\widehat{G\!R}}^{\text{dec}}}=$~0.432 (0.297)}
                & \raisebox{6pt}{\makecell{\includegraphics[height=64pt]{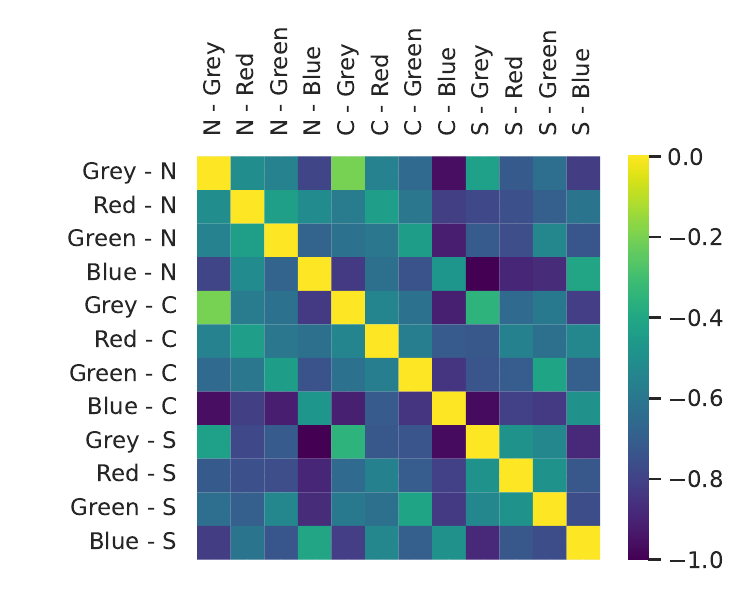}}}
                & \raisebox{1pt}{\makecell{\includegraphics[height=52pt]{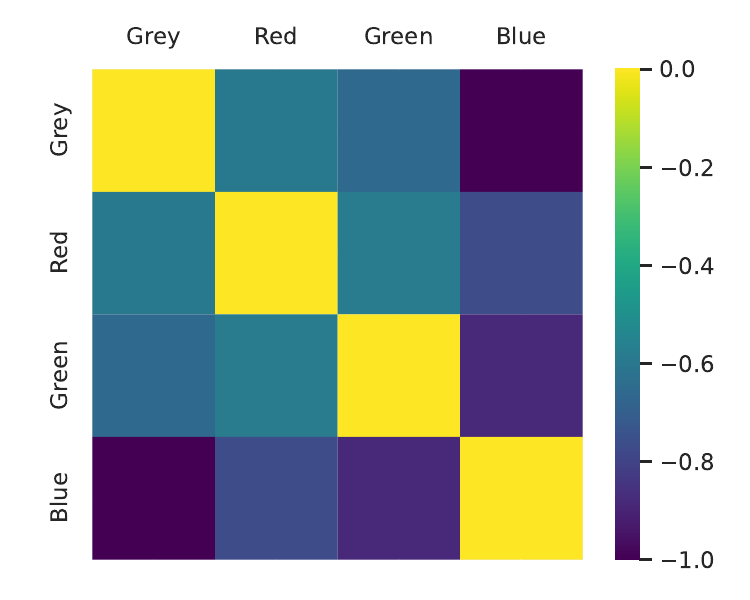}}}
                & \raisebox{1pt}{\makecell{\includegraphics[height=52pt]{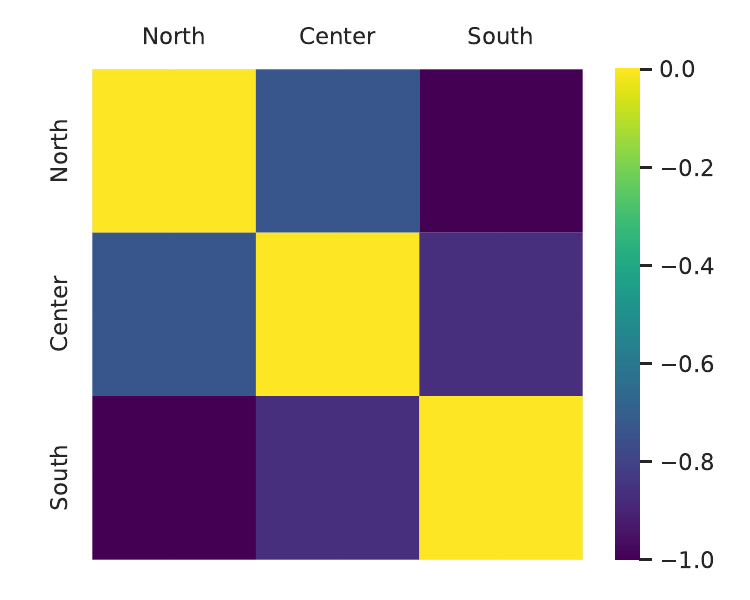}}} \\
            \makecell[l]{Identity\\(\bf\textit{Over-Specific})}
                & \makecell{\includegraphics[clip,trim={16px 16px 16px 16px},height=48pt]{figures/gen-minigrid/filter-id.png}}
                & \makecell{
                    $\mathmakebox[24pt][r]{R}=$~0.002 (0.003) \\
                    $\mathmakebox[24pt][r]{\smash{\widehat{G\!R}}}=$~0.911 (0.171) \\
                    $\mathmakebox[24pt][r]{\smash{\widehat{G\!R}}^{\text{rec}}}=$~0.264 (0.388) \\
                    $\mathmakebox[24pt][r]{\smash{\widehat{G\!R}}^{\text{dec}}}=$~0.647 (0.365)}
                & \raisebox{6pt}{\makecell{\includegraphics[height=64pt]{figures/gen-minigrid/id.pdf}}}
                & \raisebox{1pt}{\makecell{\includegraphics[height=52pt]{figures/gen-minigrid/id-cols.pdf}}}
                & \raisebox{1pt}{\makecell{\includegraphics[height=52pt]{figures/gen-minigrid/id-door.pdf}}} \\[-5pt]
            \makecell[l]{HideColors\\(\bf\textit{Ideal})}
                & \makecell{\includegraphics[clip,trim={16px 16px 16px 16px},height=48pt]{figures/gen-minigrid/filter-hidecols.png}}
                & \makecell{
                    $\mathmakebox[24pt][r]{R}=$~0.000 (0.000) \\
                    $\mathmakebox[24pt][r]{\smash{\widehat{G\!R}}}=$~0.000 (0.000) \\
                    $\mathmakebox[24pt][r]{\smash{\widehat{G\!R}}{}^{\text{rec}}}=$~0.000 (0.000) \\
                    $\mathmakebox[24pt][r]{\smash{\widehat{G\!R}}{}^{\text{dec}}}=$~0.000 (0.000)}
                & \raisebox{6pt}{\makecell{\includegraphics[height=64pt]{figures/gen-minigrid/hidecols.pdf}}}
                & \raisebox{1pt}{\makecell{\includegraphics[height=52pt]{figures/gen-minigrid/hidecols-cols.pdf}}}
                & \raisebox{1pt}{\makecell{\includegraphics[height=52pt]{figures/gen-minigrid/hidecols-door.pdf}}} \\[-5pt]
            \makecell[l]{HideDoor\\(\bf\textit{Under-Specific})}
                & \makecell{\includegraphics[clip,trim={16px 16px 16px 16px},height=48pt]{figures/gen-minigrid/filter-hidedoor.png}}
                & \makecell{
                    $\mathmakebox[24pt][r]{R}=$~0.070 (0.126) \\
                    $\mathmakebox[24pt][r]{\smash{\widehat{G\!R}}}=$~0.837 (0.157) \\
                    $\mathmakebox[24pt][r]{\smash{\widehat{G\!R}}{}^{\text{rec}}}=$~0.607 (0.227) \\
                    $\mathmakebox[24pt][r]{\smash{\widehat{G\!R}}{}^{\text{dec}}}=$~0.230 (0.217)}
                & \raisebox{6pt}{\makecell{\includegraphics[height=64pt]{figures/gen-minigrid/hidedoor.pdf}}}
                & \raisebox{1pt}{\makecell{\includegraphics[height=52pt]{figures/gen-minigrid/hidedoor-cols.pdf}}}
                & \raisebox{1pt}{\makecell{\includegraphics[height=52pt]{figures/gen-minigrid/hidedoor-door.pdf}}} \\
            \makecell[l]{Blind\\(\bf\textit{Extremely}\\\hspace{3pt}\bf\textit{Under-Specific})}
                & \makecell{\includegraphics[clip,trim={16px 16px 16px 16px},height=48pt]{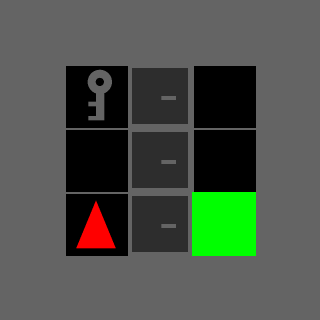}}
                & \makecell{
                    $\mathmakebox[24pt][r]{R}=$~0.279 (0.121) \\
                    $\mathmakebox[24pt][r]{\smash{\widehat{G\!R}}}=$~0.279 (0.121) \\
                    $\mathmakebox[24pt][r]{\smash{\widehat{G\!R}}^{\text{rec}}}=$~0.256 (0.101) \\
                    $\mathmakebox[24pt][r]{\smash{\widehat{G\!R}}^{\text{dec}}}=$~0.023 (0.028)}
                & \raisebox{6pt}{\makecell{\includegraphics[height=64pt]{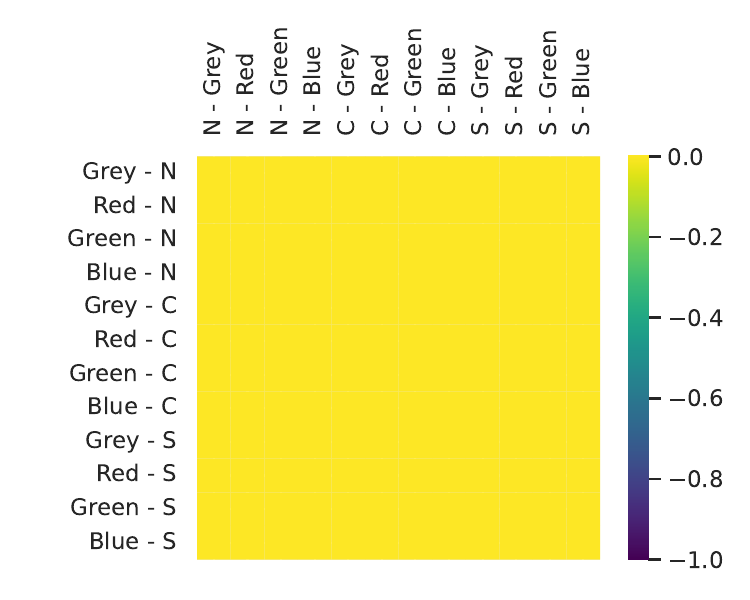}}}
                & \raisebox{1pt}{\makecell{\includegraphics[height=52pt]{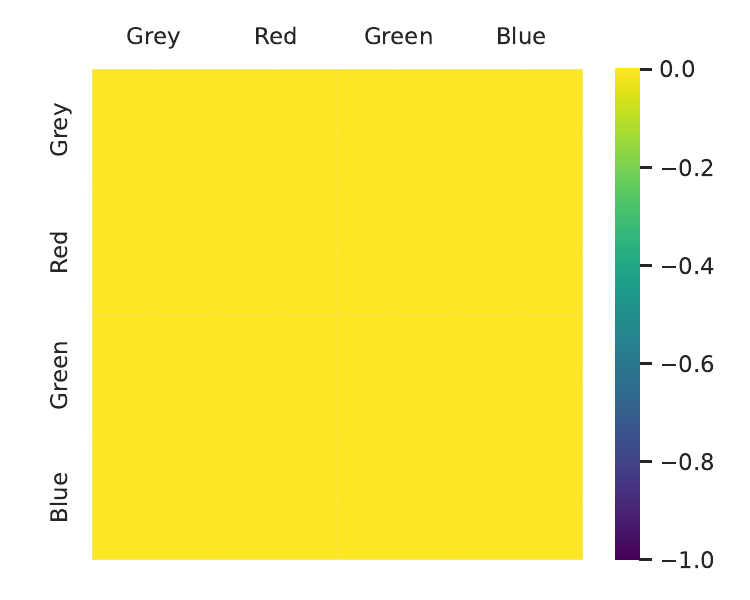}}}
                & \raisebox{1pt}{\makecell{\includegraphics[height=52pt]{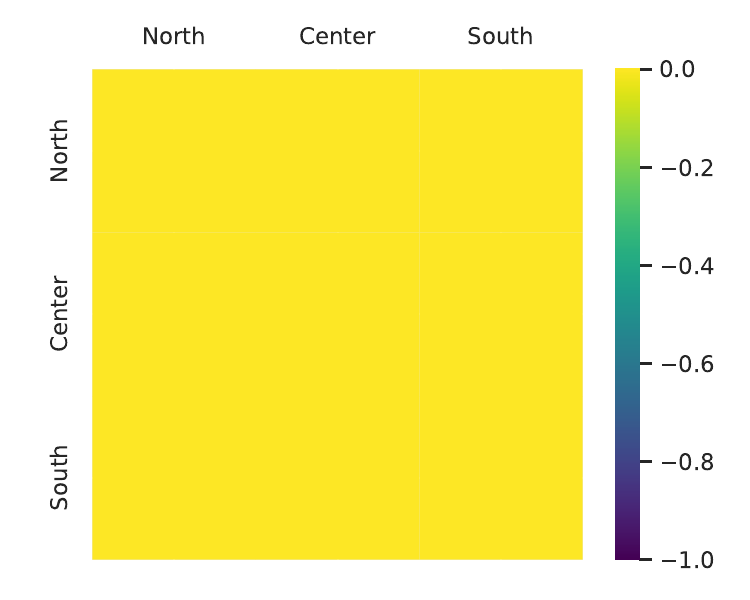}}} \\
            \bottomrule
        \end{tabular}}
\end{table}

\section{Details of the Experiments}
\label{sec:appendix}

We used a host machine with a 12-core AMD Ryzen 9 7900X CPU and two Nvidia RTXA5000 (Ampere) GPUs running Red Hat Enterprise Linux 9 to spin up Ubuntu 22.04 LTS machines for training, allocating 32 GB RAM, 2 vCPUs and 1 RTXA5000 GPU per virtual machine. Using one virtual machine, the training for all experiments have taken approximately five days.

\subsection{Experiments in Section~\ref{sec:contribution1b}}
\label{sec:appendix-first}

\paragraph{Minigrid}
The recognition policy $\rho_{\text{CNN}}$ is a sequence of: a convolutional  layer of channel size $16$ and kernel size $2$ with ReLU activations, a maximum pooling layer of kernel size $2$, a convolutional  layer of channel size $32$ and kernel size $2$ with ReLU activations, and a convolutional layer of channel size $64$ and kernel size $2$ with ReLU activations. The decision policy $\pi_{\text{MLP}}$ is a fully-connected network with a hidden layer of size $64$ and tanh activations. The PPO algorithm requires a value head in addition to the decision policy / action head; we let the value head copy the architecture of the decision policy and share the same recognition policy / encoder with the decision policy.

We consider the ``\texttt{MiniGrid-SimpleCrossingS11N5-v0}'' environment within the Minigrid package \citep{maxime2023minigrid}. Agents are trained in $64$ parallel environments for $100,\!000$ total time steps. We use the default hyper-parameters for the PPO algorithm in the skrl package \citep{serrano2023skrl}, except we collect $4096$ total transitions in between each policy update, set the entropy loss scale as $0.01$ and the value loss scale as $0.5$, and activate the running standard scalers for input images and output values. Agents are evaluated by rolling out $1024$ full episodes. We repeat all our experiments five times to obtain 1-sigma error bars.

\looseness-1
\paragraph{Pong}
The recognition policy $\rho_{\text{CNN}}$ is a sequence of: a convolutional layer of channel size $32$, kernel size $8$, and stride length $4$ with ReLU activations, a convolutional layer of channel size $64$, kernel size $4$, and stride length $2$ with ReLU activations, a convolutional layer of channel size $64$, kernel size $3$, and stride length $1$ with ReLU activations, and a fully-connected layer of size $512$ with ReLU activations \citep[known as the ``NatureCNN''][]{mnih2015human}. The decision policy $\pi_{\text{MLP}}$ is a fully-connected network with two hidden layers of size $64$ and tanh activations. The PPO algorithm requires a value head in addition to the decision policy / action head; we let the value head copy the architecture of the decision policy and share the same recognition policy / encoder with the decision policy.

We consider the ``\texttt{ALE/Pong-v5}'' environment within the gymnasium package \citep{towers_gymnasium_2023}. Normally, the game outputs $160$-by-$240$ full-color images. We crop and down-sample the game field to reduce these to $80$-by-$80$ images, and also average all color channels to obtain grayscale images. Finally, we stack the four most recent images to account for memoryless agents. Agents are again trained in $64$ parallel environments for $100,\!000$ total time steps. We use the default hyper-parameters for the PPO algorithm in the skrl package, except we collect $4096$ total transitions in between each policy update, set the entropy loss scale as $0.01$ and the value loss scale as $0.5$, and activate the running standard scalers for input images and output values as well as the KL adaptive learning rate scheduler. Agents are evaluated by rolling out $256$ full episodes. We repeat all our experiments five times to obtain 1-sigma error bars.

\subsection{Experiments in Section~\ref{sec:contribution3a}}

We consider the same policy architecture as in Appendix~\ref{sec:appendix-first} given for \textit{Minigrid}. However this time, we consider the custom maze configurations given in Figure~\ref{fig:minigrid-environments}. Agents are trained in $64$ parallel environments for $50,\!000$ total time steps. We use the default hyper-parameters for the PPO algorithm in the skrl package\footnote{%
The number of learning epochs during each update is $8$, the number of minibatches during each learning epoch is $2$, the discount factor ($\gamma$) is $0.99$, the coefficient $\lambda$ used for generalized advantage estimation is $0.95$, the learning rate is $0.001$ (an Adam optimizer is used), the clipping coefficient for the norm of the gradients is $0.5$, and the clipping coefficient for computing the clipped surrogate objective is $0.2$.
}, except we collect $4096$ total transitions in between each policy update, set the entropy loss scale as $0.01$ and the value loss scale as $0.5$, and activate the running standard scalers for output values. Agents are evaluated by rolling out $1024$ full episodes. We repeat all our experiments five times to obtain 1-sigma error bars.

\subsection{Experiments in Section~\ref{sec:contribution3b}}

We consider the same policy architecture as in Appendix~\ref{sec:appendix-first} given for \textit{Pong}. However this time, we consider a modified version of ``\texttt{ALE/Pong-v5}''. First, we set the repeated action probability as $0$ to make the game deterministic and we reset the game as soon as the first point is scored to keep the episode lengths short. In order to obtain unseen initial states, we take between $1$ and $21$ null actions (inclusive, picked uniformly at random), where taking no null action corresponds to the initial state that is seen during training.

Images outputted by the game are processed as in Appendix~\ref{sec:appendix-first} to obtain $80$-by-$80$ images. We extend these to $82$-by-$82$ images by adding a black border at the top and at the two sides. Then, we add a frame counter at the top in the form of a progress bar: When the game is on the $n$-th frame, the first $n$ pixels of the top two pixel rows are set to white (otherwise they are left black). We generate $82$-by-$82$ images as distractions and concatenate these with the original image to obtain the final images that are then fed forward to the agents.
The generating function for these distractions is as such \citep[similar to][]{song2020observational}: We start with the four-dimensional game state (consisting of the xy-coordinates of the ball and the y-coordinates of the paddles). We then pass this matrix through a neural network consisting of a fully-connected layer that brings the dimensions up to a $7$-by-$7$ matrix followed by three transposed convolutions of kernel sizes $3,4,8$ and stride lengths $1,2,4$ respectively \citep[we do not apply any activation functions, mimicking][]{song2020observational}. As we have already described in Section~\ref{sec:contribution3b}, we randomly pick from one of two generating functions at the start of each episode. We obtain different functions to pick from through different initializations of the network of transposed convolutions. Crucially, in order to not lose the information contained within the original game state, we initialize the layers within these networks with orthogonal weights.

Finally, agents are trained in $64$ parallel environments for $50,\!000$ total time steps. We use the default hyper-parameters for the PPO algorithm in the skrl package, except we collect $4096$ total transitions in between each policy update, set the entropy loss scale as $0.01$ and the value loss scale as $0.5$, and activate the running standard scalers for input images and output values as well as the KL adaptive learning rate scheduler. Agents are evaluated by rolling out $256$ full episodes. We repeat all our experiments five times to obtain 1-sigma error bars.

\subsection{Experiments in Section~\ref{sec:contribution3c}}

We consider the same setup as in Section~\ref{sec:contribution3a}. Only difference is that we vary the channel size of the last convolutional layer. During these experiments, especially for models with very small channel sizes, PPO tends to output unstable solutions. To account for the increased variation, we repeat all our experiments ten times instead of five to obtain 1-sigma error bars.

\end{document}